\pdfoutput=1

\documentclass[10pt,twocolumn,letterpaper]{article}
\usepackage[pdftex]{graphicx}

\usepackage{cvpr}              
\usepackage{multicol}
\usepackage{multirow}
\usepackage[nolist]{acronym}

\definecolor{cvprblue}{rgb}{0.21,0.49,0.74}
\usepackage[pagebackref,breaklinks,colorlinks,allcolors=cvprblue]{hyperref}

\usepackage[accsupp]{axessibility}

\begin{acronym}
    \acro{NN}[NN]{Neural Network}
    \acro{CNN}[CNN]{Convolutional Neural Network}
    \acro{RL}[RL]{Reinforcement Learning}
    \acro{IRL}[IRL]{Inverse Reinforcement Learning}
    \acro{IL}[IL]{Imitation Learning}
    \acro{BC}[BC]{Behavioral Cloning}
    \acro{ER}[ER]{Event Rate}
    \acro{MDP}[MDP]{Markov Decision Process}
    \acro{VO}[VO]{Visual Odometry}
    \acro{ROI}[ROI]{Region of Interest}
    \acro{APE}[APE]{Absolute Pose Error}
\end{acronym}

\def\paperID{25} 
\def\confName{CVPR}
\def\confYear{2025}

\title{BiasBench: A reproducible benchmark for tuning the biases of event cameras}

\author{Andreas Ziegler\thanks{equal contribution\\This research was partially funded by Sony AI.}\qquad David Joseph$^{*}$\qquad Thomas Gossard$^{*}$\qquad Emil Moldovan\qquad Andreas Zell\\
Cognitive Systems Group, University of Tübingen, Germany\\
{\tt\small \{andreas.ziegler,david.joseph,thomas.gossard\}@uni-tuebingen.de}
}

\begin{document}
\maketitle
\begin{abstract}
Event-based cameras are bio-inspired sensors that detect light changes asynchronously for each pixel.
They are increasingly used in fields like computer vision and robotics because of several advantages over traditional frame-based cameras, such as high temporal resolution, low latency, and high dynamic range.
As with any camera, the output's quality depends on how well the camera's settings, called biases for event-based cameras, are configured.
While frame-based cameras have advanced automatic configuration algorithms, there are very few such tools for tuning these biases.
A systematic testing framework would require observing the same scene with different biases, which is tricky since event cameras only generate events when there is movement.
Event simulators exist, but since biases heavily depend on the electrical circuit and the pixel design, available simulators are not well suited for bias tuning.
To allow reproducibility, we present BiasBench, a novel event dataset containing multiple scenes with settings sampled in a grid-like pattern.
We present three different scenes, each with a quality metric of the downstream application.
Additionally, we present a novel, RL-based method to facilitate online bias adjustments.
%
%
%
\end{abstract}

\section*{Additional Resources}
Additional resources are available at: \url{https://cogsys-tuebingen.github.io/biasbench}

\section{Introduction}\label{sec:intro}

\begin{figure}[t!]
  \centering
  \includegraphics[width=1.0\linewidth]{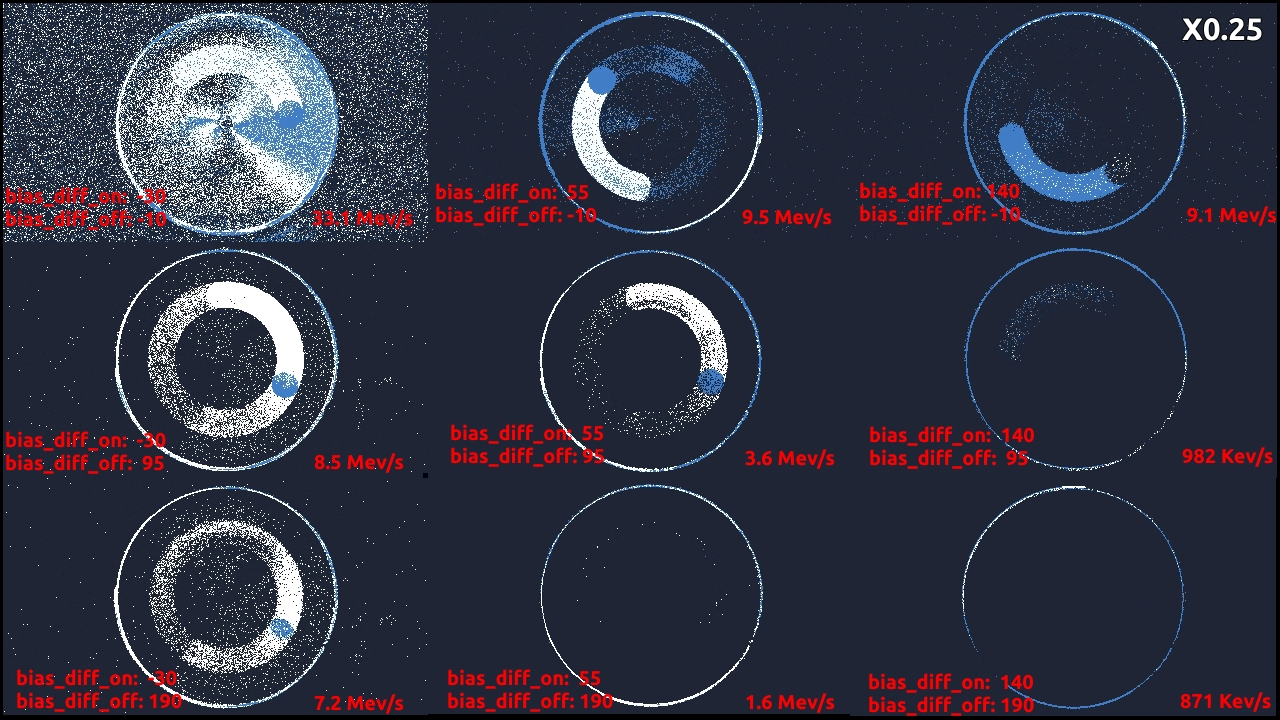}
  \caption{
    Accumulated event frames of the spinning disk setup of our event dataset with different \textit{bias\_diff\_on} and \textit{bias\_diff\_off}.
    As can be seen, different bias settings heavily influence the signal-to-noise ratio.    
  }
  \label{fig:eye-catcher}
\end{figure}

The human eye operates differently from a traditional, frame-based camera.
Its retinal receptors respond to changes in light within the environment, converting these changes into signals that are then transmitted to the brain.
The eye can detect as many as a thousand light changes per second while ignoring the mostly static background.
Event-based cameras, inspired by the human visual system, address several limitations of frame-based cameras, including low temporal resolution, high latency, limited dynamic range, and data redundancy.
These bio-inspired sensors operate asynchronously, with each pixel independently detecting light changes and generating \textit{events} that indicate whether luminosity has increased or decreased.
This mechanism results in superior temporal resolution, allowing event-based cameras to capture fast movements with significantly less motion blur.
Since only changes in light intensity are recorded, static parts of the image do not produce \textit{events}.
This makes event-based cameras very useful in many computer vision and robotics applications where a high temporal resolution, low latency, and a high dynamic range are needed.

Event-based cameras are configured through five parameters, commonly referred to as \textit{biases}, which govern event detection thresholds and signal filtering.
These five \textit{biases} are the positive and negative thresholds for the event generation, the cut-off frequencies for a low- and high-pass filter, and the refractory period, which defines the dead time after a pixel generated an event, defined as the tuple $(\textit{bias\_diff\_on}, \textit{bias\_diff\_off}, \textit{bias\_fo}, \textit{bias\_hpf}, \textit{bias\_refr})$.
There is no universal bias configuration: they depend on the downstream task.
Event-based cameras have been used for a wide range of applications, e.g., object tracking~\cite{Perot2020neurips}, image segmentation~\cite{Alonso2019cvprw, Zhou2021tnnls}, odometry~\cite{HidalgoCarrio2022cvpr}.
They are particularly effective at capturing high-frequency phenomena~\cite{Baldini2024isprs}, as well as scenes in both low-light conditions~\cite{Vidal2018ral} and bright sunlight~\cite{Gehrig2021ral}, provided the biases are appropriately selected.
This implies that the biases are tailored to the specific task.
The full potential of event-based cameras can, therefore, only be realized when they are properly configured.
A favorable choice of biases can result in a low noise and high contrast output, which is beneficial for the downstream application.

This raises the question of how to optimally control the biases of an event-based camera.
In order to develop algorithms for bias tuning we need to be able to evaluate the effect of different settings. For this, event data from the same scene with different bias settings is needed.
Furthermore, if we want to compare different algorithms, the event data should be reproducible.
Recreating the same real-world recording setup in different research labs seems difficult to achieve.
As for other applications, event simulators could be considered as an option.
However, the sim2real gap could cause problems, especially since the biases depend very much on the electric circuit and the pixel design of the event camera, which are not accurately modeled in available event simulators.
Therefore, we present a novel dataset that contains multiple scenes with a grid sampling of bias settings. A snapshot is shown in~\cref{fig:eye-catcher}.
Instead of using a real event camera to develop bias-tuning algorithms, our novel event dataset can be used.
After the algorithm proposes a set of biases, the corresponding dataset recording can be used instead of changing the biases of an actual event camera.
This allows a reproducible event stream without the need of a carefully designed calibration setup, speeding up the development of bias-tuning algorithms.
Next to the event recordings with different sets of biases, we also provide a quality metric for the event streams of each of the three setups.
Additionally, we introduce an initial baseline algorithm for tuning the two most commonly used biases, the on and off thresholds, leveraging our dataset to enable online bias optimization.

When tuning the biases, we are sequentially adjusting the biases to get the best event stream for our downstream application.
This process can be formulated as a \ac{MDP}, defined by the tuple $(\mathcal{S}, \mathcal{A}, \mathcal{P}, r)$, where:
\begin{description}
	\setlength\itemsep{0.025em}
	\item[$\bullet$ $\mathcal{S}$]: observed event stream under the current bias settings.
	\item[$\bullet$ $\mathcal{A}$]: the set of possible bias modifications that influence the event generation process.
	\item[$\bullet$ $\mathcal{P}: \mathcal{S} \times \mathcal{A} \rightarrow \mathcal{S}$]: the probabilistic evolution of the event stream in response to bias changes.
	\item[$\bullet$ $r: \mathcal{S} \times \mathcal{A} \rightarrow \mathbb{R}$]: the reward for a given bias change.
\end{description}
However, both $\mathcal{P}$ and $r$ are inaccessible in practice.
Modeling $\mathcal{P}$ is infeasible for two reasons: (1) Event cameras capture brightness changes rather than absolute intensity, making it difficult to isolate bias effects from scene dynamics.
(2) The event camera's response to biases is highly nonlinear and interdependent, lacking a precise physical model.
Consequently, model-based \ac{RL} is not applicable.
As for $r$, it could be defined via properties like event sparsity, noise reduction, or task performance.
While \ac{ER} can quantify sparsity, real-time noise estimation remains unavailable, and event stream quality depends on the task.
To bypass the need for a cost function, we propose to instead use \acf{BC} to adjust camera biases.

\textbf{Contributions} of this work are as follows:
\begin{itemize}
    \item We present a novel event dataset for bias tuning, containing multiple scenes covering a grid of bias settings
    \item For each scene, we provide a quality metric and, therefore, a quality measure for each bias configuration
    \item We present a \acf{BC} pipeline as a baseline algorithm that is able to tune the two most common biases of event cameras online
    \item We make the recorded dataset, the metric, and the baseline algorithm publicly available
\end{itemize}
\section{Related Work}\label{sec:related_work}

We structure the related work as follows:
First, we discuss bias settings and the best practices for manual tuning in~\cref{subsec:biases}. 
Then, we review approaches for automatic bias tuning in~\cref{subsec:bias_tuning}.
Finally, we cover key aspects of \ac{RL} relevant to this study in~\cref{subsec:rl}.

\subsection{Bias Settings}\label{subsec:biases}

A first step towards automatic bias tuning is understanding the biases and their effect on the event stream.
Several papers meticulously document the behavior of the event-based camera.
In~\cite{Graa2023cvprw}, the behavior of the DVS pixels for different bias parameters is described through extensive experimentation and measurements.
In addition, the authors propose a new way to interpret events as a combination of signal, noise, and leakage.
An outline of the manual bias optimization process is given, and general indications are proposed for different scenarios.
This, and several other papers~\cite{Gossard2024cvprw,Graca2023iiss}, have described the process of manually finding optimal biases.
The most widely used metric in the literature for rating an event camera's output is the \acf{ER}~\cite{Delbruck2021cvprw,Nair2024iros}.
However, in cases where the event camera is used for object detection, another way of rating the output is by measuring its sharpness, a method proposed in~\cite{SefidgarDilmaghani2023icmv}.
This is achieved by calculating the average gradient of the frame.
The authors conclude that this method shows promise but highlights the potential advantages of a metric that operates directly on the event stream without requiring frame generation.

In all the mentioned work, tuning the biases has to be done manually.
Other works have simply used the default settings of the camera~\cite{Amir2017cvpr} or do not mention them~\cite{Gehrig21ral}.
Our work provides the necessary dataset to develop algorithms that tune the biases automatically.
In addition, we also present a first baseline algorithm.

\subsection{Automatic Bias Tuning}\label{subsec:bias_tuning}

Comparatively, little has been done in the field of automatic bias tuning, given the plethora of research around event-based computer vision.
Multiple papers~\cite{Graa2023cvprw,Gossard2024cvprw,Nair2024iros} dealing with the problem of finding optimal bias settings have resorted to measuring and rating the output for a range of different bias settings and then manually changing these biases.

In~\cite{Delbruck2021cvprw}, the authors presented a closed-loop control method, representing a first step towards automatic bias tuning, using a combination of three fixed-step feedback controllers.
These feedback controllers take measurements of the \ac{ER} and noise and regulate them using all five bias parameters.
The noise is calculated using a denoising algorithm and measuring how many events are removed.
An optimal range is defined for the \ac{ER} and the noise, and a simple algorithm changes the bias values if the two measurements leave their optimal ranges.
The \ac{ER} is regulated by increasing the thresholds if it is too high and decreasing them if it is too low.
The \ac{ER} is also limited by increasing the refractory period.
The noise is controlled by adjusting the bandwidth.
One limitation is that achieving the optimal bias setting requires multiple steps.
Additionally, each threshold adjustment can trigger a burst of events, potentially leading to a short-term decline in the quality of the event stream.
Our proposed baseline does not require multiple steps, as it suggests the bias change necessary to reach an optimal bias in one step.

Another approach, and the most advanced work on the topic of which we are aware, builds on the same idea but uses a two-part algorithm~\cite{Nair2024iros}.
The algorithm controlling the biases uses two methods: fast adaptation, which changes the refractory period in order to limit the \ac{ER}, and slow adaptation, which consists of changing the thresholds and adjusting the low-pass filter if fast adaptation is not enough to limit the \ac{ER}.
This split into two adaptations is necessary because only the refractory period does not cause visual artifacts when changed, which is problematic for the visual place recognition task presented in~\cite{Nair2024iros}.
This means that the other three bias parameters should not be changed too often.
For downstream applications such as visual(-inertial) odometry, SLAM, or object detection, too much noise could lead to missing objects or losing track of the trajectory.
While this approach is more advanced, it only considers the \ac{ER} as a metric for the quality of the event stream.
On the contrary, in our work, we rely on expert demonstrations so that the pipeline mimics an expert and does not only rely on the \ac{ER}.

\subsection{Reinforcement Learning}\label{subsec:rl}

Instead of manually designing a control algorithm, \ac{RL} has proven to be successful in learning complex control policies given enough demonstrations~\cite{Silver2017nature,Jumper2021nature}.
In a more similar task, like adjusting the exposure time of a frame-based camera, \ac{RL} was also successfully applied~\cite{Lee2024cvpr}.
For \ac{RL} algorithms to converge to something meaningful, a valid reward function is needed.
This even led to its own research directions with reward engineering~\cite{Singh2019rss} and reward shaping~\cite{Laud2004}.

Unfortunately, we do not have access to a metric capable of evaluating the quality of the events compared to frame-based cameras, which makes vanilla \ac{RL} not applicable.
As is evident from the literature on bias tuning for event-based cameras, there is no universal metric to measure the quality of an event stream.
Many works use the \ac{ER} as a quality indicator, but this is too short-sighted.
This is applicable to certain tasks, but depending on the scene, the \ac{ER} might vary greatly and is therefore not transferable to other problems.
Instead of going the route of reward engineering, we rely on \acf{IL}, a subgroup of \ac{RL}.
The reward function is either learned from data, which is called \ac{IRL}, or the behavior is learned directly from expert demonstrations in a supervised fashion, which is called \acf{BC}.

In a first work~\cite{Abbeel2004icml}, the authors introduced an apprenticeship learning algorithm that uses demonstrations from an expert to learn via \acf{IRL}.
\ac{IRL} was further improved by employing the principle of maximum entropy in~\cite{Ziebart2008ncai}.
It is further extended by using a Fully Convolutional Neural Network (FCNN) for the reward instead of a linear mapping of features, which allows for more complex feature mappings~\cite{Wulfmeier2015arxiv}.
Data-driven \ac{RL} has also been used in robotics applications, e.g., to control a robot arm with images as input~\cite{Singh2019rss}.

In this work, we developed an \ac{BC} policy for bias tuning, which serves as a baseline.
Specifically, we use TD3+BC~\cite{Fujimoto2021neurips}, a simple approach to offline \ac{RL} where TD3~\cite{Scott2018icml} has been changed in two ways:
A weighted behavior cloning loss is added to the policy update, and the states are normalized.

\section{Dataset}\label{sec:dataset}

As previously mentioned, to develop algorithms for automatic bias tuning, event data from the same scene with different bias settings is needed.
Since there is no standard approach to tuning the biases on event cameras yet, there is also no standard setup to do so.
While the number of possible scenes is unlimited, everyone could build their own setup.
We believe reproducible data in the form of a benchmark can help the community to make comparisons.

Although there are multiple event simulators available these days~\cite{Rebecq2018corl,Hu2021v2e,Joubert2021event,Ziegler2023icra}, none of them model all the bias settings accurately, which is essential given that the biases are very much dependent on the electric circuit and pixel design of an event camera.
Therefore, we present a dataset containing multiple setups with various bias settings, each with two scenes to be used for training/development and test.
Changing the biases of the camera will be simulated by changing the file being read, so as to match the new biases. 
For each setup, we provide a quality metric and a quality measurement for each recorded bias configuration.
This dataset is intended to serve the community as a benchmark and enable the development of bias-tuning algorithms.

All recordings were captured using Prophesee EVK4 event cameras.
For these event cameras, the biases are expressed as relative offsets, with the default values being $0$.
The spinning dot and blinking LED datasets were recorded with an 8mm lens, while the robot arm dataset was recorded with a 6mm lens.
The focus was adjusted using the Metavision Focus Adjustment tool\footnote{\url{ttps://docs.prophesee.ai/stable/samples/modules/calibration/focus.html}}, and the aperture was set to $f/4$.

We recorded two scenes of the same setup with the same lighting conditions so that any differences between the recordings are caused by the bias settings and not by changes in the scene.
To allow for the development of model-based or learning-based algorithms for bias tuning, we need setups for which we can define a quality metric.
Thus, we built three different setups: a spinning dot, an LED board consisting of LEDs with different blinking frequencies and waveforms, and an event camera attached to a robot arm, performing a trajectory as an \ac{VO} example.

\subsection{Spinning Dot}\label{subsec:ds_spinning_dot}

\begin{figure}[t!]
    \centering
    \begin{subfigure}[b]{0.31\linewidth}
            \includegraphics[width=\linewidth]{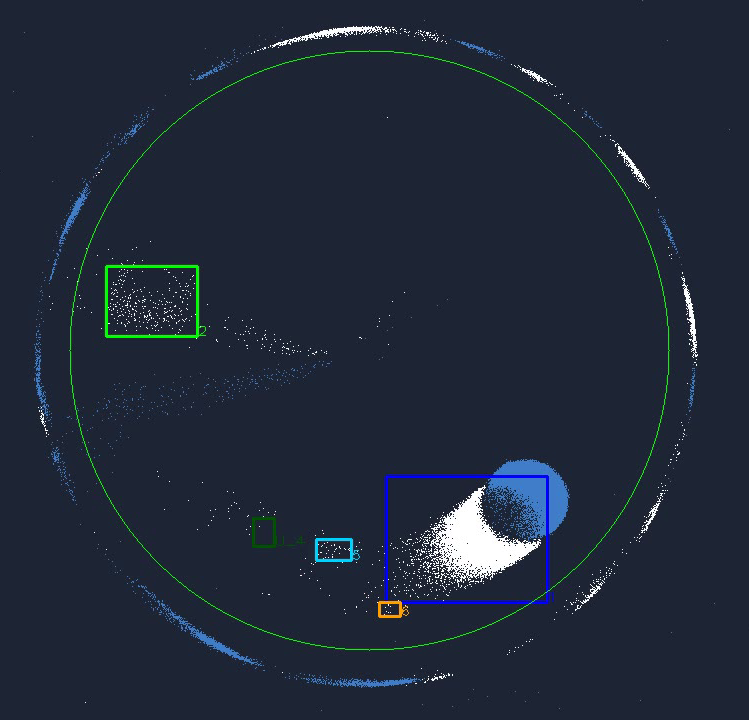}
    \end{subfigure}
    \hfill
    \begin{subfigure}[b]{0.31\linewidth}
            \includegraphics[width=\linewidth]{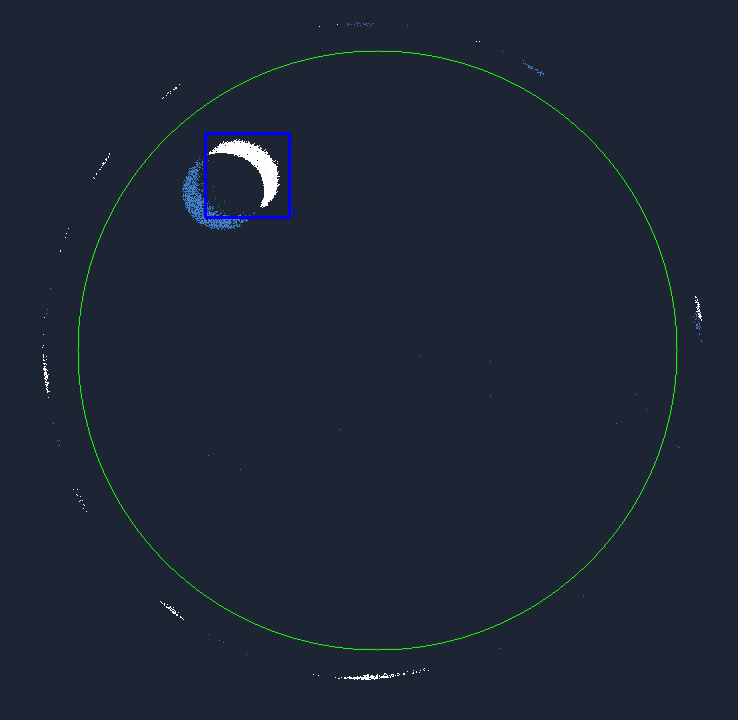}
    \end{subfigure}
    \hfill
    \begin{subfigure}[b]{0.31\linewidth}
            \includegraphics[width=\linewidth]{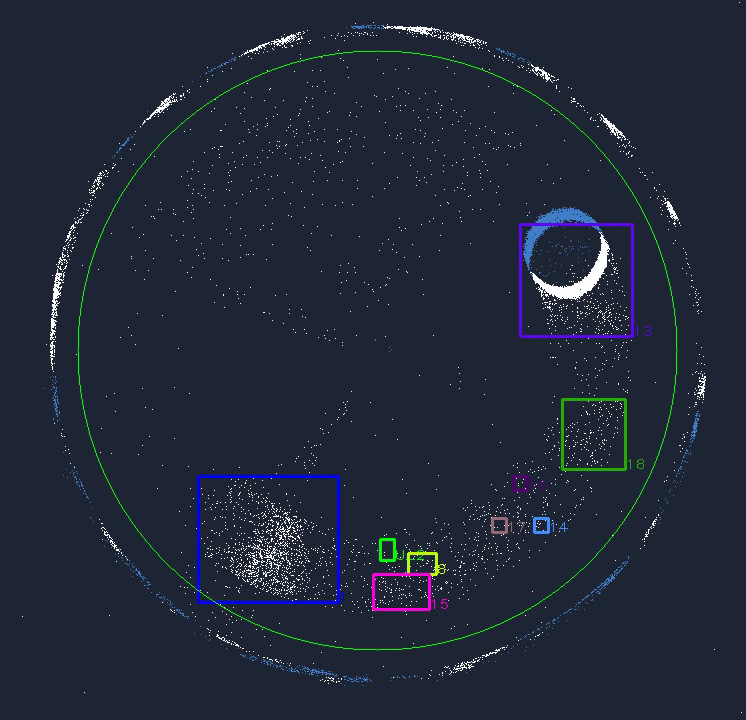}
    \end{subfigure}
    \caption{
            The spatter tracker from the Metavision SDK for different event data with the bias settings (left) $(20, -10, -35, 24, 126)$, (middle) $(20, 25, 10, 0, -20)$, and (right) $(-10, 95, 10, 0, 53)$.
            }
    \label{fig:dot_tracker}
\end{figure}

The spinning dot has become a famous example to explain the working principles of an event camera.
We brought this example alive by attaching a grey or black dot to a spinning disk.

We spin the dot at $10$ rps and film it with the event camera $0.4$m away from the dot spinner.
The setup is shown in~\cref{fig:spinning_dot_setup} in the supplementary material.
Using the event camera, we captured $38,880$ recordings, each with a grey and a black dot, with different bias settings.
Each recording is one second long.
We also wait a bit after each bias update to ignore the event burst, which can occur when changing the bias parameters.
The light source comprises LED panels without pulse modulation, resulting in a constant brightness at an experiment level of around $1500$ lumen.

For the bias settings, we choose an equidistant selection covering the whole range of the biases, listed in~\cref{tab:spinning_dot_biases}.
More details can be found in~\cref{subsec:sm_spinning_dot} in the supplementary material.
\begin{table}[t!]
  \centering
  \begin{tabular}{cccc}
  \hline
  Bias & Start & End & Number of biases \\
  \hline
  \hline
  \textit{bias\_diff\_on} & -30 & 130 & 18\\
  \textit{bias\_diff\_off} & -10 & 190 & 18\\
  \textit{bias\_fo} & -35 & 55 & 5\\
  \textit{bias\_hpf} & 0 & 120 & 6 \\
  \textit{bias\_refr} & -20 & 200 & 4 \\
  \hline
  \end{tabular}
  \caption{
        Bias values for the spinning dot setup.
        We choose an equidistant bias selection.
        }
  \label{tab:spinning_dot_biases}
\end{table}
To obtain a quality metric for the spinning dot, we used the spatter tracker from the Metavision SDK\footnote{\url{https://docs.prophesee.ai/stable/samples/modules/analytics/tracking\_spatter\_py.html\#chapter-samples-analytics-tracking-spatter-python}}, shown in~\cref{fig:dot_tracker}.
The accumulation time was set to $1$ms, and the tracking was limited to within the spinning disk to ignore events generated by the edge of the disk.

In this work, we introduce three distinct metrics for evaluating the quality of the tracking, taking into account the influence of the biases of event cameras.
The first metric is the tracklet fragmentation $\textit{TF}$, which quantifies the number of distinct tracklets corresponding to the spinning dot. 
Ideally, this value should be $1$, indicating continuous tracking without interruption or fragmentation.
The second metric is the tracking length $\textit{TL}$, representing the total duration for which the dot is being tracked. 
This includes periods when the dot is tracked under different tracking identifiers (IDs).
A longer tracking length suggests more consistent tracking, though varying tracklet IDs may indicate tracking challenges.
Additionally, we record the total number of different tracklets $N_{tracklets}$ detected during the tracking process.
In an ideal scenario, this number should be $1$, reflecting that only the spinning dot is being tracked.
Any additional tracklets are typically due to noise or false positives, such as detected spatters, which we aim to mitigate.
We consider biases valid when $\textit{TL}>0.75$.
Under this criterion, 44\% of the settings result in successful dot tracking for the black dot.

\subsection{Blinking LED Board}\label{subsec:ds_led_board}

\begin{figure}[t!]
    \centering
    \begin{subfigure}[b]{0.48\linewidth}
        \centering
        \includegraphics[width=\linewidth]{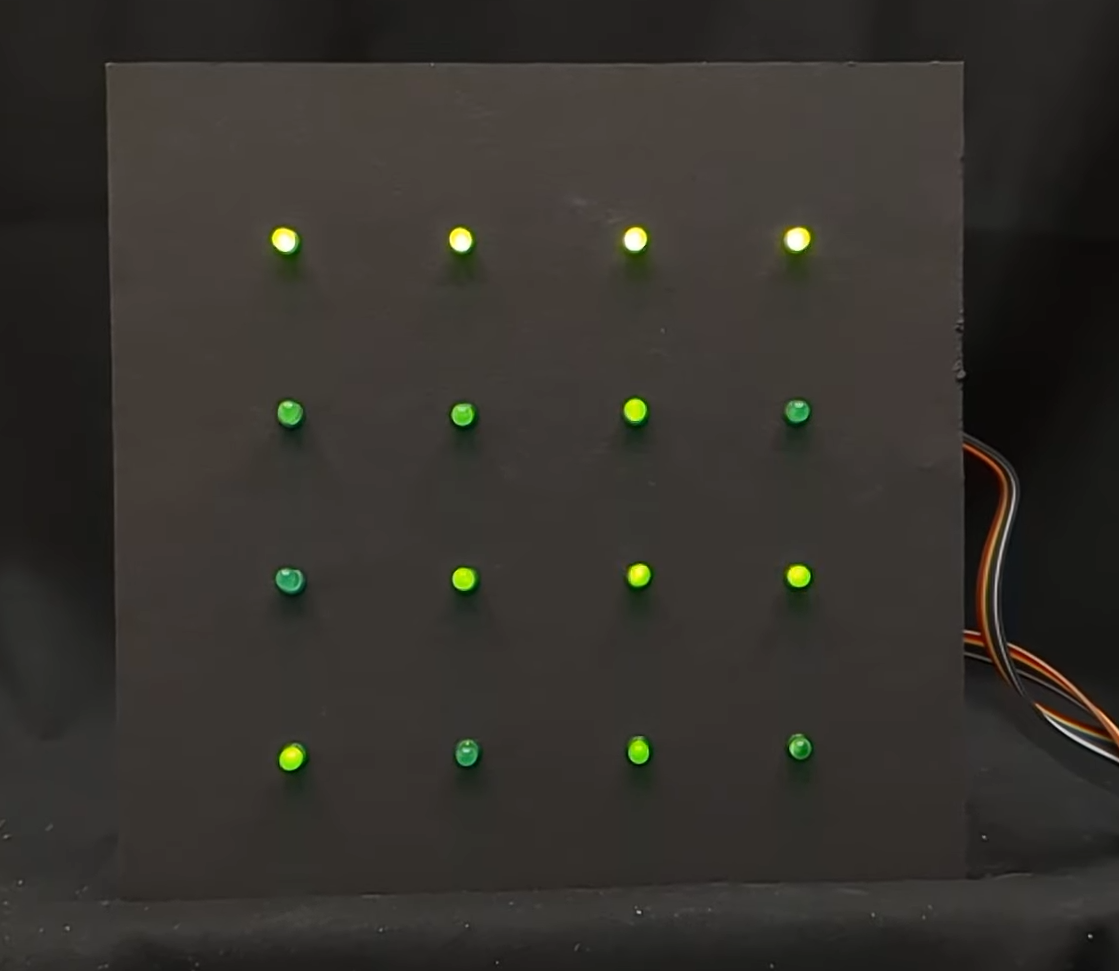}
        \caption{Blinking LED board}
        \label{fig:blinking_led_board}
    \end{subfigure}
    \hfill
    \begin{subfigure}[b]{0.48\linewidth}
        \centering
        \includegraphics[width=\linewidth]{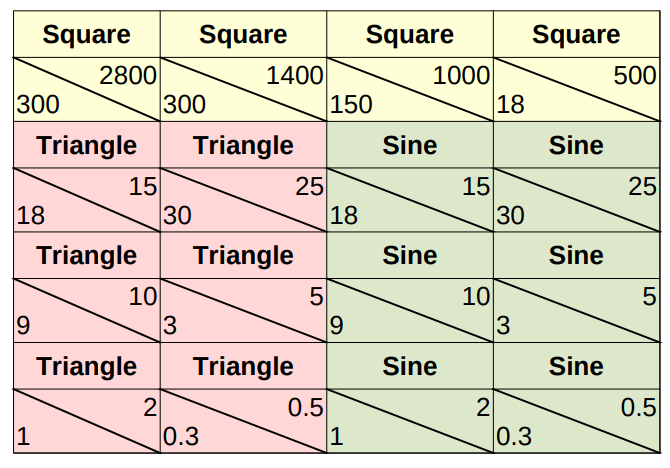}
        \caption{
                Waveform for each corresponding LED, bottom left values are the test set frequencies, and top right are the train set frequencies.}
        \label{fig:led_frequencies}
    \end{subfigure}
    \caption{Blinking LED setup}
    \label{fig:led_board_frequencies}
\end{figure}

To see the influence of the low-pass and high-pass filter, controlled by the biases \textit{bias\_fo} and \textit{bias\_hpf}, we designed an LED board, shown in~\cref{fig:blinking_led_board}.
We used a $4 \times 4$ LED grid, each emitting different waveforms to introduce varying light intensity patterns, as visualized in~\cref{fig:led_frequencies}.

The four LEDs in the top row emitted a square waveform with a $0.5$ duty cycle, each operating at a different frequency.
The six LEDs on the lower left followed a sine wave with six distinct frequencies, while the six on the lower right emitted a triangular waveform with matching frequencies.
This setup was designed to ensure diverse light intensity variations.
We set the frequencies to different values to distinguish the train and test set.
These frequencies are described in \Cref{fig:led_frequencies}.

For the bias settings, we choose an equidistant selection covering the whole range of the biases, listed in~\cref{tab:led_board_biases}.
This gave us a total of $30,967$ recordings.
More details can be found in~\cref{subsec:sm_led_board} in the supplementary material.
\begin{table}[t!]
  \centering
  \begin{tabular}{cccc}
  \hline
  Bias & Start & End & Number of biases \\
  \hline
  \hline
  \textit{bias\_diff\_on} & -80 & 120 & 11\\
  \textit{bias\_diff\_off} & -30 & 170 & 11\\
  \textit{bias\_fo} & -29 & 48 & 8 \\
  \textit{bias\_hpf} & 4 & 116 & 8 \\
  \textit{bias\_refr} & -15 & 225 & 4\\
  \hline
  \end{tabular}
  \caption{
        Bias values for the LED board setup.
        We choose an equidistant bias selection.}
  \label{tab:led_board_biases}
\end{table}

As a quality metric, we estimated the frequencies of the LEDs by fitting a cosine function to the difference in positive and negative events within the region associated with the LED.
With this estimated frequency and its error, $f_{est}$ and $\Delta f_{est}$ respectively, the true frequency, $f_0$, we define our metric as Relative Frequency Uncertainty (RFU)
\begin{equation}
    RFU= \frac{\left| f_{est}-f_0 \right|+\Delta f_{est}}{f}.
\end{equation}
A value of $0$ indicates a perfect frequency estimation with a negligible error, while higher values represent the relative size of misalignment and error compared to value.
We cut this at a value of $2$ to not overestimate on single bad-fitted frequencies.
We consider biases valid when every frequency could be estimated with an RFU of below $2$.
Under this criterion, $55\%$ of the settings result in a successful frequency estimation.
\begin{figure}[t!]
  \centering
  \includegraphics[width=1.0\linewidth]{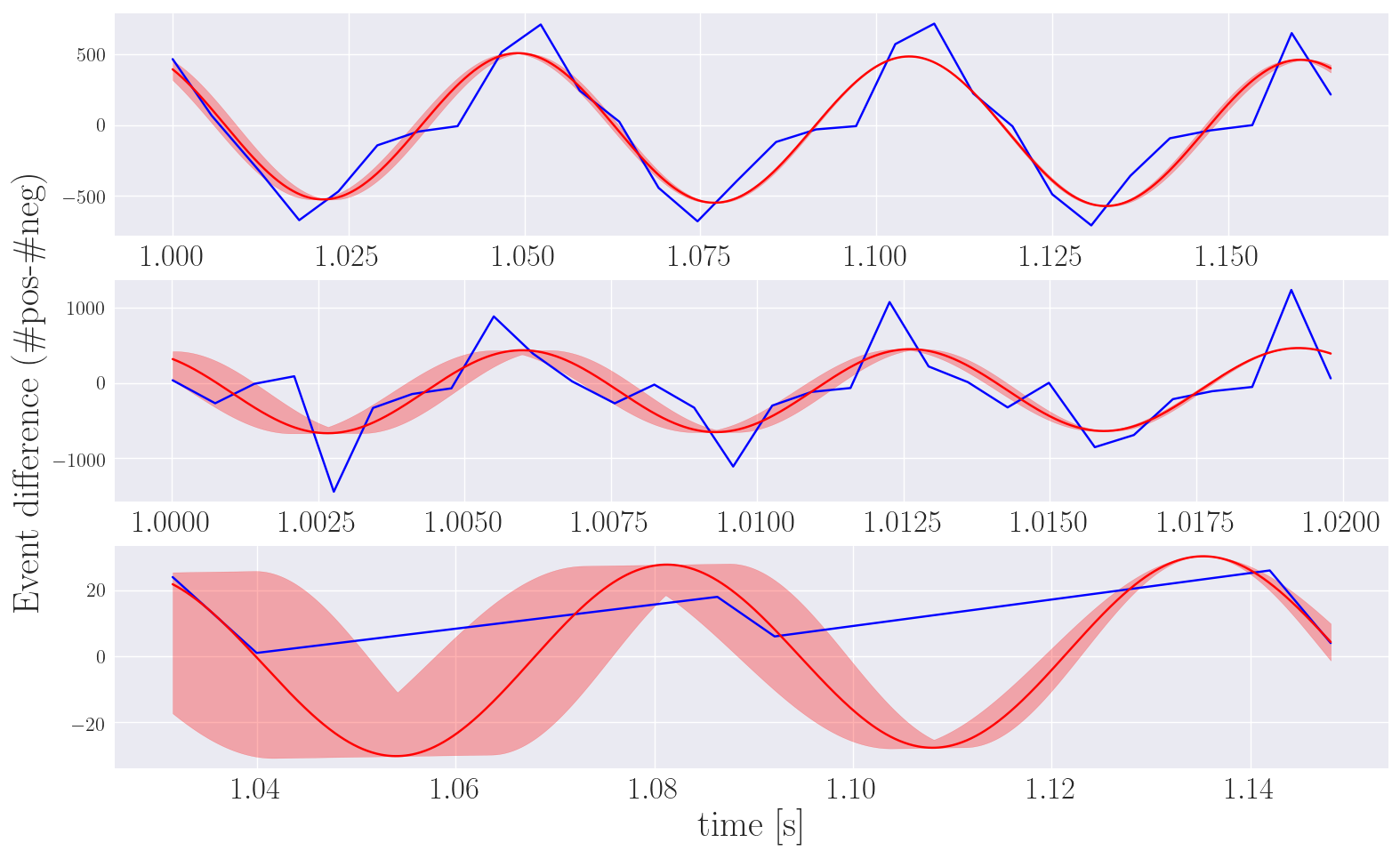}
  \caption{
    Examples of the frequency estimation for different LEDs and bias settings.
    The real event data is given {\color{blue}in blue}, the best fit {\color{red}in red}.
    The red-shaded area resembles the uncertainty of the fit from the frequency estimate.
    The resulting Relative Frequency Uncertainty is for the function on the top $0.018$, for the function in the middle $0.04$, and for the function on the bottom $0.4$.
  }
  \label{fig:freq_estimation}
\end{figure}

\subsection{Visual Odometry (VO)}\label{subsec:ds_vo}

\begin{figure}[t!]
  \centering
  \includegraphics[height=18em,width=1.0\linewidth]{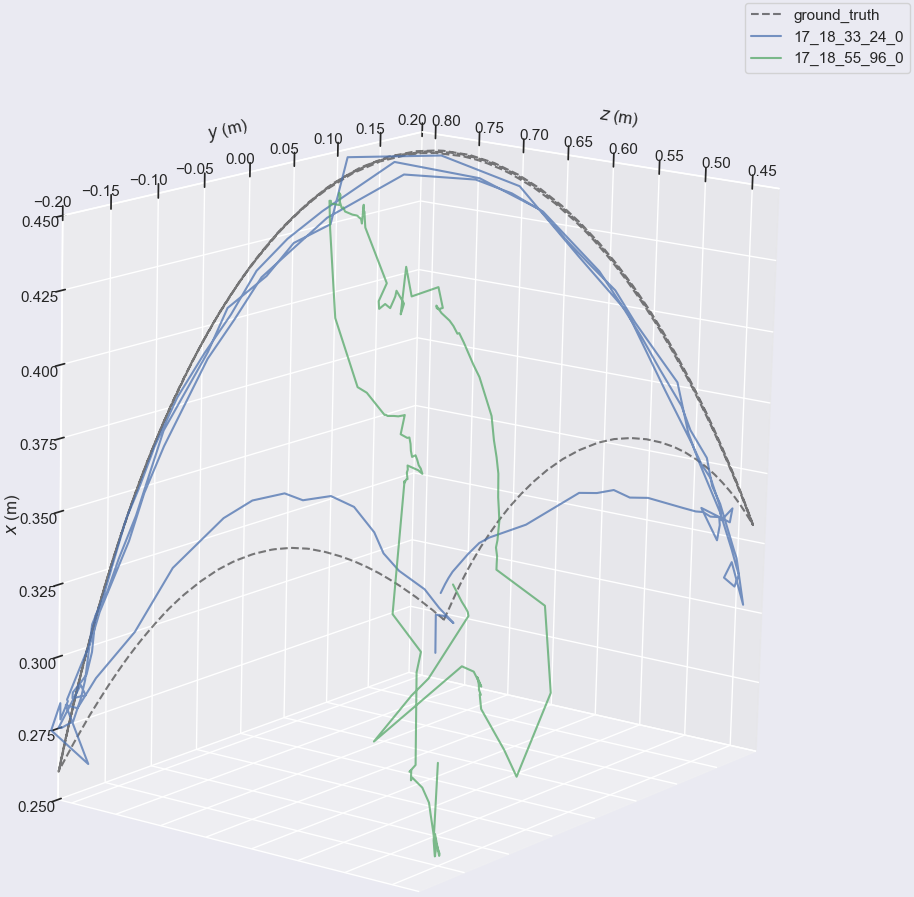}
  \caption{
    The ground truth trajectory of the end-effector of a Pandas robot arm is shown in {\color{black} in black}.
    The trajectory output from DEVO~\cite{Klenk20243dv} for the \textit{bias} settings $(17, 18, 33, 24)$ is shown in {\color{blue} in blue}.
    And the trajectory output for the \textit{bias} settings $(17, 18, 55, 96)$ is shown in {\color{green} in green}.
    Despite the similar bias values, the \ac{VO} failed in one case.
  }
  \label{fig:vo_left_right_trajectory}
\end{figure}

For a more application-oriented setup, we recorded event data intended to be used for event-based \ac{VO} pipelines.
We recorded the events for a trajectory in which the camera moved left and right in one plane for $10$s and a triangle trajectory with a duration of $11$s.
To get a repeatable motion and an accurate ground truth trajectory, we mounted an event camera to the end-effector of a Pandas robot arm, shown in~\cref{fig:vo_setup} in the supplementary material.
The first ground truth trajectory from the end-effector of the robot arm is shown in~\cref{fig:vo_left_right_trajectory}.

For the bias settings, we choose an equidistant selection, covering the whole range of the biases listed in~\cref{tab:vo_biases}.
More details can be found in~\cref{subsec:sm_vo} in the supplementary material.
\begin{table}[t!]
  \centering
  \begin{tabular}{rlll}
  \hline
  Bias & Start & End & Number of biases \\
  \hline
  \hline
  \textit{bias\_diff\_on} & -50 &140 & 15 \\
  \textit{bias\_diff\_off} & -10& 190& 15\\
  \textit{bias\_fo} & -30 & 55 & 5 \\
  \textit{bias\_hpf} & 0 & 120 & 6\\
  \textit{bias\_refr} & 0 & 0 & 1\\
  \hline
  \end{tabular}
  \caption{
        Bias values for the \ac{VO} setup.
        We choose an equidistant bias selection.}
  \label{tab:vo_biases}
\end{table}

To get a metric, we run DEVO~\cite{Klenk20243dv}, a state-of-the-art event-based \ac{VO} pipeline, on all event recordings and evaluated the resulting trajectories with evo~\cite{Grupp2017evo}.
This led to $6750$ recorded event streams per trajectory.

\begin{figure}[t!]
    \centering
    \begin{subfigure}[b]{0.48\linewidth}
        \centering
        \includegraphics[width=\linewidth]{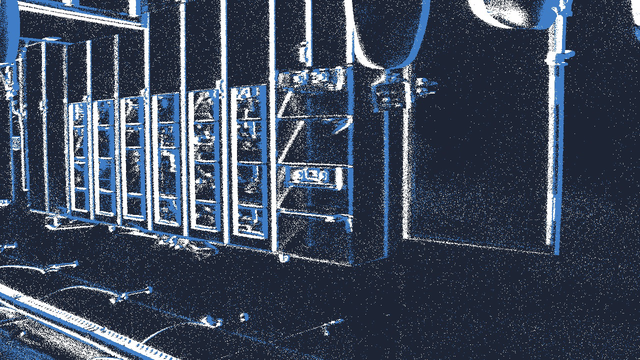}
        \captionsetup{justification=centering}
        \caption{Success \\ biases $=[17, 18, 33, 24, 0]$}
        \label{fig:success}
    \end{subfigure}
    \hfill
    \begin{subfigure}[b]{0.48\linewidth}
        \centering
        \includegraphics[width=\linewidth]{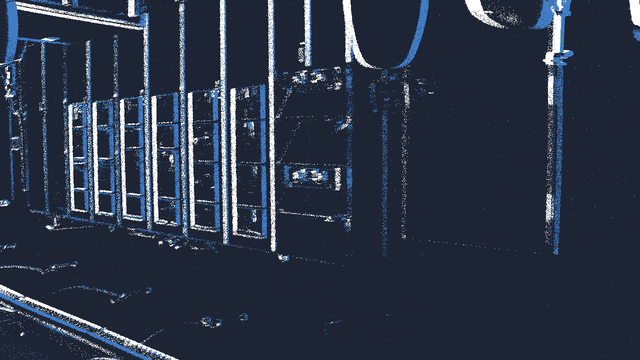}
        \captionsetup{justification=centering}
        \caption{Failure \\ biases $=[17, 18, 55, 96, 0]$}
        \label{fig:failure}
    \end{subfigure}
    \caption{Comparison of accumulated event frames of (a) a successful and (b) a failed estimated trajectory with different biases.}
    \label{fig:devo_comparison}
\end{figure}

In \Cref{fig:devo_comparison}, we present accumulated event frames for two different bias configurations.
Although the event frames appear visually similar, DEVO successfully inferred the trajectory in one case but failed in the other, as shown in~\cref{fig:vo_left_right_trajectory}.
This underscores the fact that even when the event stream looks comparable to human observers, its effectiveness in downstream tasks can vary significantly.

\begin{figure}[t!]
  \centering
  \includegraphics[width=1.0\linewidth]{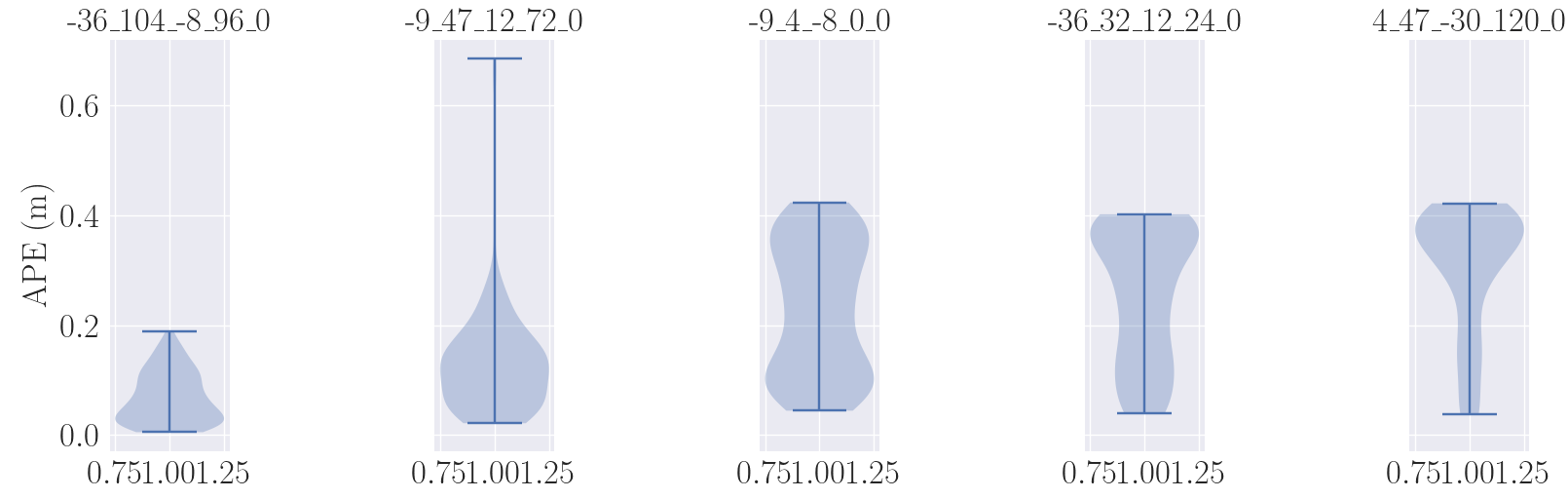}
  \caption{
        The \acf{APE} plot of five trajectories, each with different bias settings, estimated by DEVO~\cite{Klenk20243dv}.
        These five trajectories are picked to have an equidistant error between each other.
        }
  \label{fig:vo_left_right_errors}
\end{figure}

From all the recordings of the first trajectory, we visualize the \ac{APE} of five selected ones in~\cref{fig:vo_left_right_errors}.
These recordings are chosen such that their errors are evenly spaced across the range of observed errors.

From~\cref{fig:vo_left_right_errors}, it can be seen that only the recording with the bias settings $(-9, 4, -8, 0, 0)$ led to a reasonable \ac{APE}.
This shows the importance of bias tuning in order for a downstream application to give reasonable results.

We consider biases valid when $\textit{APE} < 0.13$.
Under this criterion, $10.05\%$ of the settings result in a successful trajectory estimate.

\section{A Baseline}\label{sec:baseline}

In addition to the presented dataset and to showcase how our dataset can be used, we developed a \ac{BC} based policy baseline to tune the two most commonly used biases, \textit{bias\_diff\_on} and \textit{bias\_diff\_off} with the other biases set to default values.

\subsection{Method}\label{subsec:method}

As previously mentioned, we framed bias optimization as a \ac{MDP} defined by the tuple $(\mathcal{S}, \mathcal{A}, \mathcal{P}, r)$.
Since formulating a good reward function is complicated and there is not yet a cost function the literature agrees on, we choose an \ac{IL} based approach.
Therefore, we leverage expert demonstrations using our dataset and employ a \ac{BC} algorithm to learn a policy for adjusting the two threshold biases, \textit{bias\_diff\_on} and \textit{bias\_diff\_off}.
We used the recorded event streams from our novel dataset and manually generated expert demonstrations by specifying the biases we estimated were optimal.

The input of the model is a compressed latent space of the event stream that we get from a feature extraction pipeline and an expert demonstration.
The feature extraction is done in multiple steps.
We start by accumulating \textit{ON}-events and \textit{OFF}-events for an accumulation time of $8$ms, each resulting in one channel.
These two channels have the resolution of the used event camera $(1280\times720$ pixels).
Due to hot pixels, there are pixels with some events that are orders of magnitude higher than the rest of the pixels.
This will lead to a decrease in sensitivity in the nondefect pixels after normalization.
Therefore, we do not normalize the input by dividing with the maximal value. We linearly scaled the 90\% of data with the lowest event count to 0.9 while mapping the remaining 10\% to the range from 0.9 to 1.
This way, we do not ignore hot pixels but do not let them dominate the input data.

\begin{figure}[t!]
  \centering
  \includegraphics[trim={0.7cm 0.2cm 0.9cm 0.2cm},clip,width=1.0\linewidth]{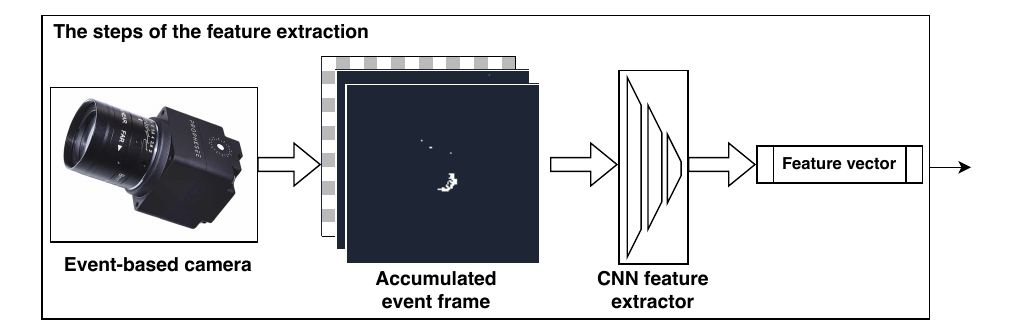}
  \caption{
    The steps of the feature extraction from the event stream to a vector of floats.
    We use ResNet50~\cite{He2016cvpr} as a feature extractor.}
  \label{fig:feature_extraction}
\end{figure}

The next step is applying a pre-trained ResNet~\cite{He2016cvpr} as a feature extraction algorithm on the normalized input.
To conform with the input dimensions of ResNet, we added an empty third channel.
ResNet uses an input resolution of $224\times 224$, whereas our accumulated event frames have a resolution of $1280\times 720$ pixels.
Thus, we divide the histograms into sub-frames and apply the feature extraction on each of these sub-frames.
For each subframe, we get a feature vector with a size of $1000$.
These stacked feature vectors can then be used as input for the \ac{BC} learning algorithm.
All the steps in the feature extraction are sketched in~\cref{fig:feature_extraction}.

Next to the extracted features from the event data, the second input is an expert demonstration.
The demonstrations need to be formatted like the actions, so the expert must choose the absolute bias settings that seem best for each recording or change the biases needed to reach this setting.
Other work~\cite{Lee2024cvpr} found that relative changes perform better than absolute values, so we have decided to also use relative bias changes.
For the expert demonstration, we choose optimal bias settings where the \ac{ER} is low, indicating a low level of noise, while the spinning dot is clearly visible.

For the \ac{BC} algorithm we adapted the TD3+BC algorithm~\cite{Fujimoto2021neurips} from its implementation in the online library d3rlpy\footnote{\url{https://github.com/takuseno/d3rlpy}}~\cite{Seno2022jmlr}.
This adaptation includes adjusting the input to accept our features and calculating the loss using the Euclidean distance between the prediction and the expert demonstration.
The overall pipeline is illustrated in~\cref{fig:training_pipeline} in the supplementary material.

For training, we created $10000$ randomly distributed accumulated event frames.
As a loss, we implemented the mean Euclidean distance between predicted and expert action.
As hyperparameters, we chose a learning rate of $0.0003$, a batch size of $265$, a discount factor of $0.99$, and Adam as the optimizer.

To evaluate the capabilities of the learned \ac{BC} model, we performed the following experiments.

\subsection{Experiments}\label{subsec:experiments}

\paragraph{Influence of the training and evaluation data}

Having two datasets with the spinning dot, one with a black dot and one with a gray dot, we investigated the influence of the training and test dataset.
We speculated that training the model on the recordings of the black dot would lead to a model that generalizes far worse than the one trained on the recordings of the grey dot because the black dot is always visible even for the maximum values of the thresholds as opposed to the grey dot.
This means that the model would never encounter the state in which the dot is not visible in training, leading to it suggesting sub-optimal actions if it encounters that state.
We tested this hypothesis by performing different runs.
In one run, we trained the model on recordings of the black dot and validated it on the grey dot.
In another run, we trained the model on the grey dot and validated it on the black dot.
For the test dataset, we created $200$ accumulated event frames from the $100$ recordings of the other dot.
The results of this comparison in~\cref{tab:training_data} prove our hypothesis that the model trained on the black dot is less accurate.
\begin{table}[t!]
\begin{center}
    \begin{tabular}{cccc}
    \textbf{Model trained on} & \textbf{Model validate on} & \textbf{Test loss} \\
    \hline
    \hline
    Black dot dataset &Grey dot dataset & 0.26 \\
    \hline
    Grey dot dataset & Black dot dataset & 0.2 \\
    \hline
    \end{tabular}
    \caption{
        The final loss values with different training and test data after the training process finished.
        The loss is given in the Mean Squared Error of the normalized proposed action.}
    \label{tab:training_data}
\end{center}
\end{table}
\begin{table*}[t!]
  \begin{center}
    \begin{tabular}{ccclccc}
      \multirow{2}{*}{Initial biases} & \multicolumn{2}{c}{Average proposed bias change} &  & \multicolumn{2}{c}{Demonstration bias change} & \multirow{2}{*}{Success rate}\\
      \cline{2-3} \cline{5-6}
      & \textit{diff\_off} & \textit{diff\_on} &  & \textit{diff\_off} & \textit{diff\_on} \\
      \hline
      \hline
      \rowcolor[HTML]{EFEFEF} 
      -10, -35                                                                 & $21 \pm 10$ & $100 \pm 10$ & & 25 & 75                   & 1.0                                                                                  \\ \hline
      65, -10                                                                  & $6 \pm 5$   & $35 \pm 10$  & & -10 & 10                   & 0.8                                                                                   \\ \hline
      \rowcolor[HTML]{EFEFEF} 
      40, 40                                                                   & $10 \pm 5$  & $16 \pm 5$  & & 0 & 10                     & 1.0                                                                                  \\ \hline
    \end{tabular}
    \caption{Results of an object tracker experiment over $10$ runs.
      The initial bias settings are expressed as two numbers which represent the settings \textit{bias\_diff\_off} and \textit{bias\_diff\_on}.
      The last column expresses the success rate of how many times the model has found an optimal bias setting and was able to track the dot over the number of experiments.
    }
    \label{tab:object_tracker}
  \end{center}
\end{table*}

\paragraph{Object Tracker}

We used the same metrics, as defined in \Cref{subsec:ds_spinning_dot}, with three different initial bias pairs to test starting configurations.
As inadequate bias values, we used \textit{bias\_diff\_off} $= -10$ and \textit{bias\_diff\_on} $= -35$.
For sub-optimal bias values, we used \textit{bias\_diff\_off} $= 65$ and \textit{bias\_diff\_on} $= -10$.
And as optimal bias values \textit{bias\_diff\_off} $= 40$ and \textit{bias\_diff\_on} $= 40$.
After the bias values were set, we used our model to adjust the biases and then started the object tracker.

\begin{figure}[htb!]
  \centering
  \begin{subfigure}{0.49\linewidth}
    \includegraphics[width=\linewidth]{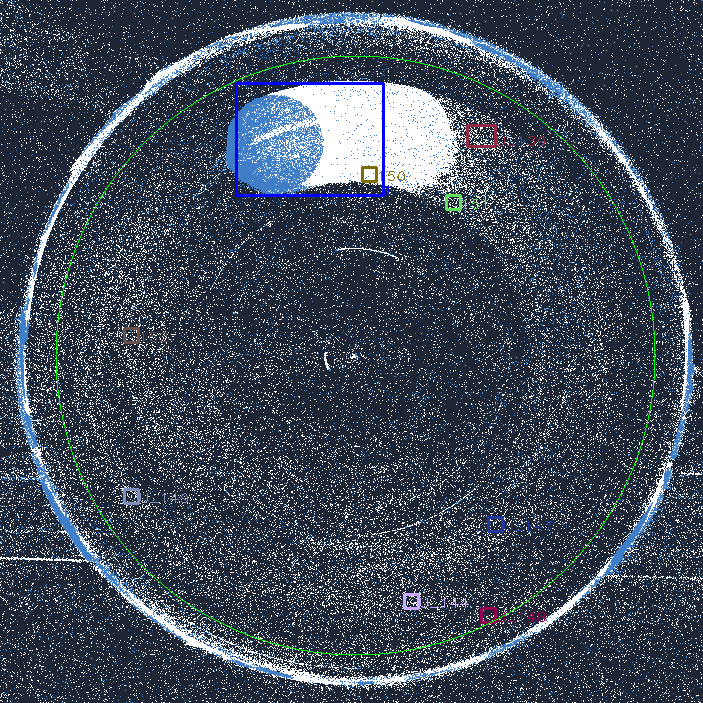}
    \caption{off = -10 and on = -35}
  \end{subfigure}
  \centering
  \begin{subfigure}{0.49\linewidth}
    \includegraphics[width=\linewidth]{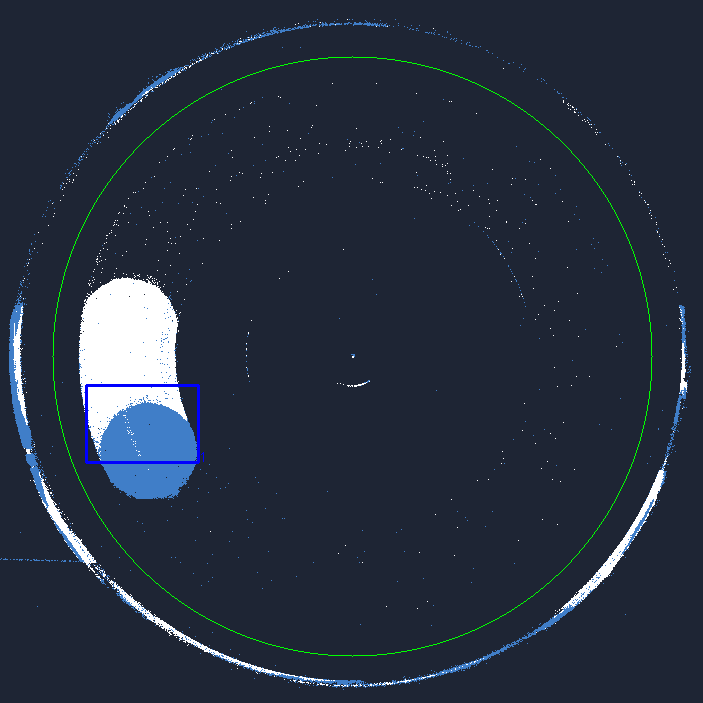}
    \caption{off = 40 and on = 40}
  \end{subfigure}
  \caption{
    Demonstration of the model proposing a bias change from (a) to (b) that allows the object tracking algorithm to accurately track the dot and not see any other objects caused by noise.}
  \label{fig:object_tracker}
\end{figure}

The results of this experiment are listed in~\cref{tab:object_tracker}, and an example of a change in bias is shown in~\cref{fig:object_tracker}.
The model struggled with large initial \textit{bias\_diff\_off} values, proposing excessive bias changes that weakened the signal, but this is uncommon in real-world scenarios where default bias values as a starting point ensure stable bias adjustments.

\paragraph{Convergence Behavior}

\begin{figure}[t!]
  \centering
  \includegraphics[width=1.0\linewidth]{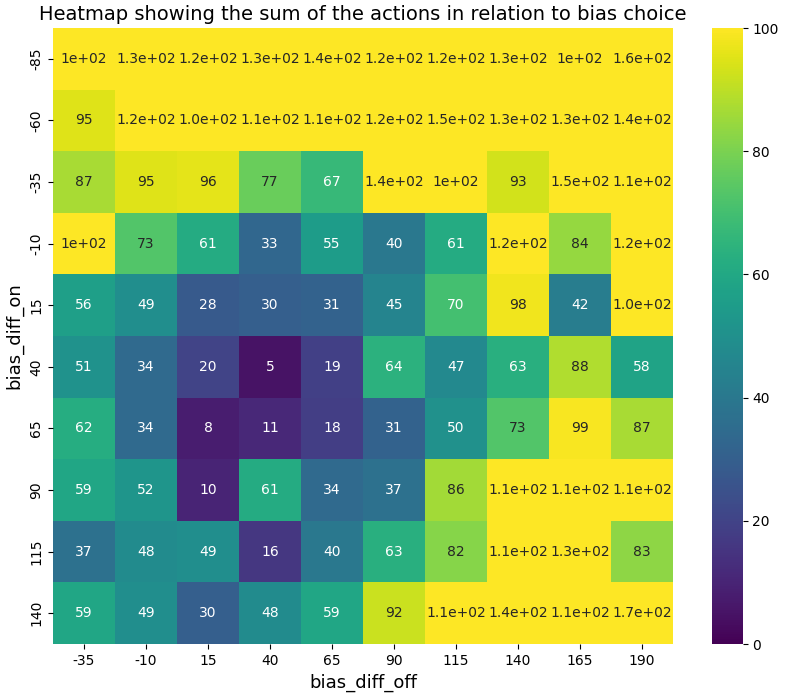}
  \caption{
    A heat map showing the sum of the relative bias changes suggested by the model for different \textit{bias\_diff\_off} and \textit{bias\_diff\_on} for the recordings of the grey dot.
    For the bias pair \textit{bias\_diff\_off} $ = 40$ and \textit{bias\_diff\_on} $ = 40$, the sum of the actions is the smallest, which resembles the optimal bias settings for this scene.
  }
  \label{fig:convergence_grey_dot}
\end{figure}

To check whether the model converges for any bias value and does not simply maximize the thresholds for every scene, we run our model for each of the $100$ grey dot recordings and report the sum of the relative bias changes of the threshold predictions.
The result as a heat map can be seen in \cref{fig:convergence_grey_dot}.
The absolute value of the actions gets smaller and smaller until they reach the setting \textit{bias\_diff\_off} $ = 40$ and \textit{bias\_diff\_on} $ = 40$, where they are closest to zero.
This indicates that the model correctly identifies this bias pair as an optimal bias setting.
Additionally, the absolute value of the actions is larger the farther away the bias settings are from this setting.

\begin{figure}[t!]
  \centering
  \includegraphics[width=1.0\linewidth]{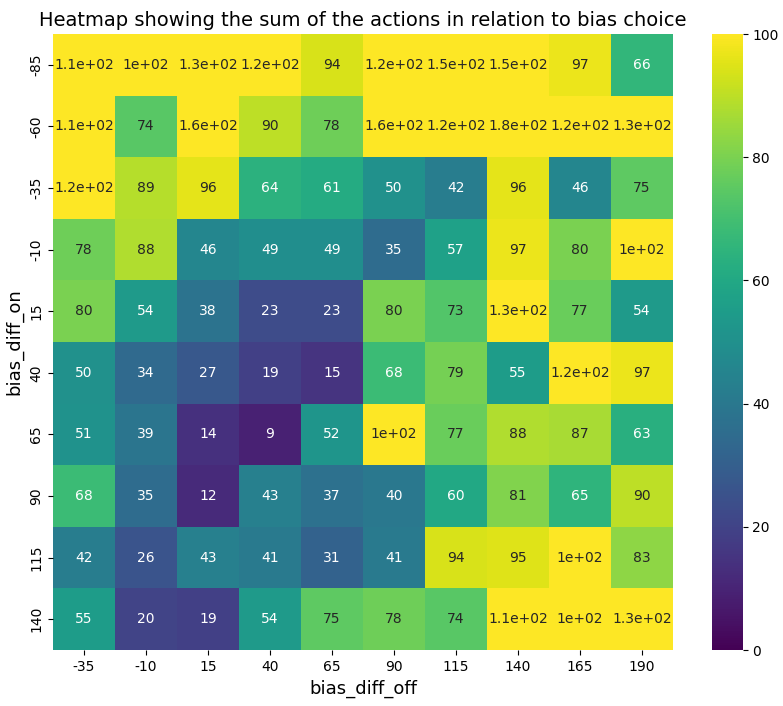}
  \caption{
    A heat map showing the relative bias changes suggested by the model for different \textit{bias\_diff\_off} and \textit{bias\_diff\_on} for the recordings of the black dot.
    For the bias pair \textit{bias\_diff\_off} $ = 40$ and \textit{bias\_diff\_on} $ = 65$, the sum of the actions is the smallest, which resembles the optimal bias settings for this scene.
  }
  \label{fig:convergence_black_dot}
\end{figure}

The same thing happens for the recordings of the black dot.
The corresponding heat map is shown in~\cref{fig:convergence_black_dot}.
We can see that the sum of the bias changes generally gets smaller the closer the biases are to the value of $40$ for \textit{bias\_diff\_off} and the value $65$ for \textit{bias\_diff\_on}.
The model converges around these values, which are different compared to the convergence values for the grey dot.
Therefore, the model recognizes the different scenes and has different predictions and optimal biases for each one.

\section{Limitations and future work}

Our approach is limited by the discrete and relatively large step sizes used during bias tuning, which may restrict fine-grained optimization.
Additionally, the metrics employed are inherently biased by the parameters of the downstream application, potentially limiting their generalizability across different tasks.
Naturally, analytical approaches such as denoising-based methods could also be used as possible metrics.

\section{Conclusion}\label{sec:conclusion}

Event-based cameras offer advantages in computer vision but require well-tuned biases for optimal performance.
To facilitate automated bias tuning, we introduced \textit{BiasBench}, a dataset of scenes captured under varying biases with corresponding quality metrics.
This dataset enables the development and evaluation of tuning algorithms, fostering a better understanding of bias selection for different tasks.
Additionally, we presented a baseline RL-based approach that learns bias tuning from expert demonstrations using \acf{BC}.

{
    \small
    \bibliographystyle{ieeenat_fullname}
    \bibliography{main}
}

\clearpage
\setcounter{page}{1}
\maketitlesupplementary

\section{Access to the Dataset}\label{sec:sm_access}

The dataset, the presented metrics as well as some additional information is accessible on our project website: \url{https://cogsys-tuebingen.github.io/biasbench/}

\section{Datasets}\label{sec:sm_datasets}

Here, we provide some additional information about the three setups of our dataset.
The spinning dot in~\cref{subsec:sm_spinning_dot}, the blinking LED board in~\cref{subsec:sm_led_board}, and the \acf{VO} in~\cref{subsec:sm_vo}.

\subsection{Spinning Dot}\label{subsec:sm_spinning_dot}

\begin{figure}[htb!]
  \centering
  \includegraphics[width=1.0\linewidth]{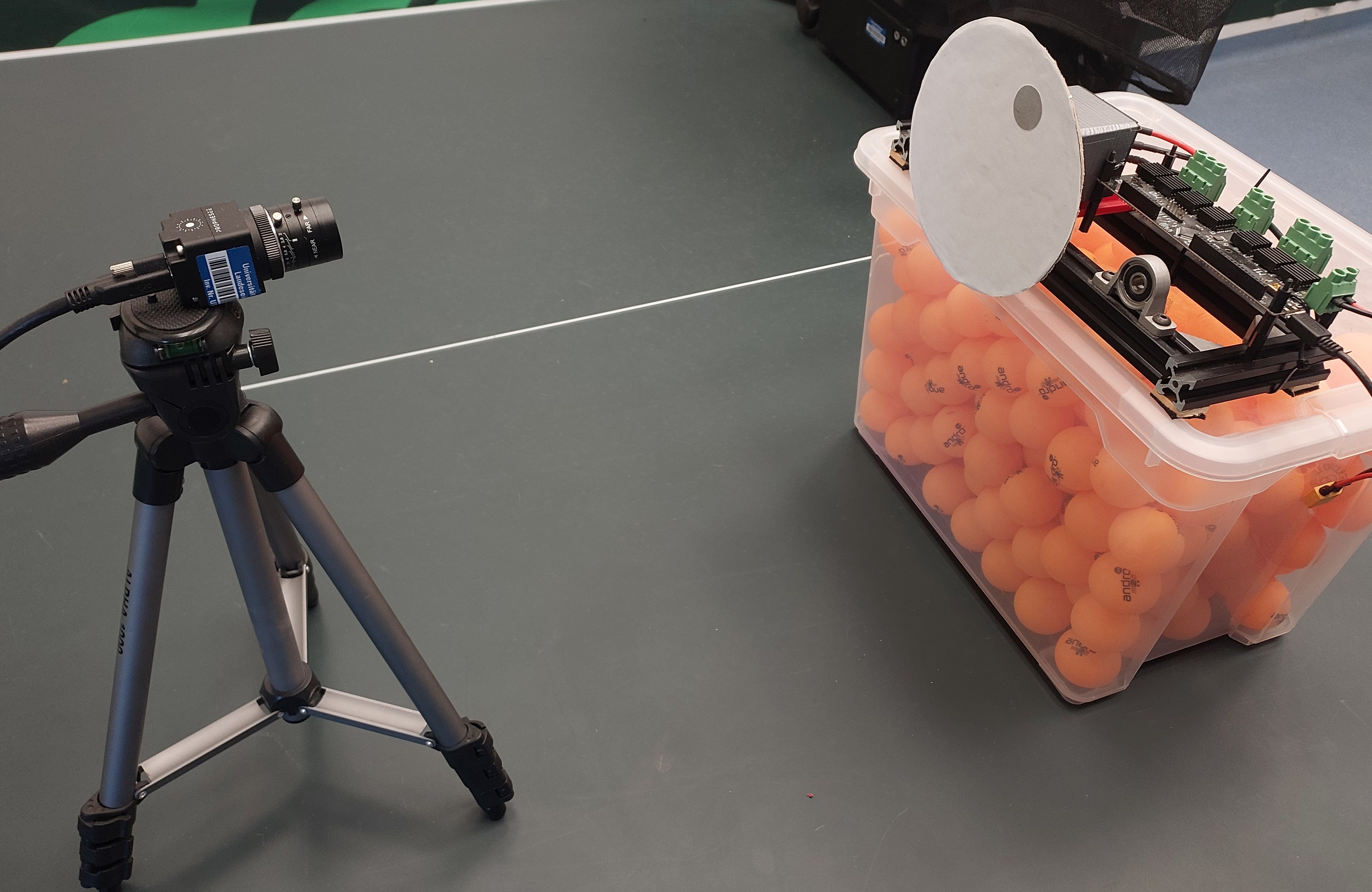}
  \caption{
    The setup of how we recorded the spinning dot consists of an EVKv4 event-based camera from Prophesee with a resolution of $1280 \times 720$ pixels and a white disk with a grey dot connected to a motor.
    The distance between the motor and the spinning disk is $40$ centimeters.
  }
  \label{fig:spinning_dot_setup}
\end{figure}

\begin{figure*}[htb!]
  \centering
  \begin{subfigure}{0.19\linewidth}
    \includegraphics[width=\linewidth]{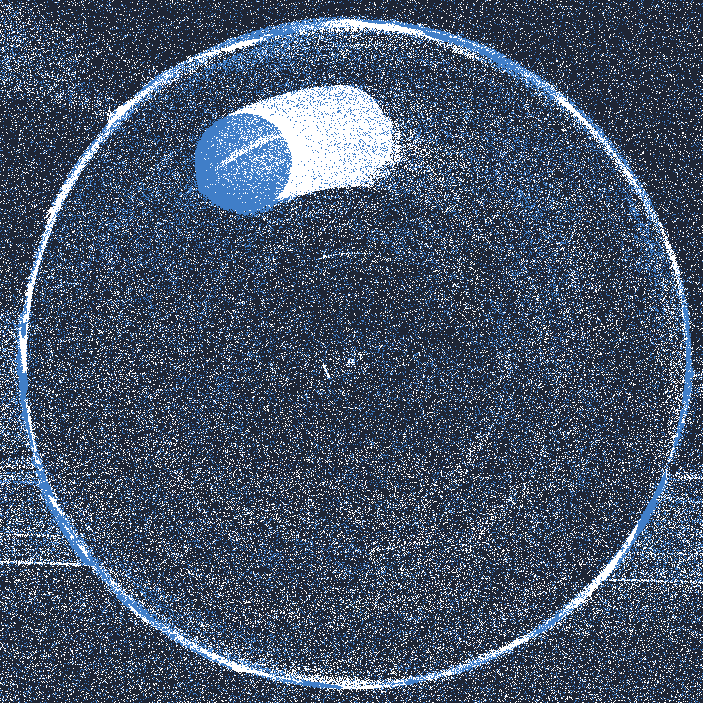}
    \caption{off = -35 and on = -35}
  \end{subfigure}
  \hfill
  \begin{subfigure}{0.19\linewidth}
    \includegraphics[width=\linewidth]{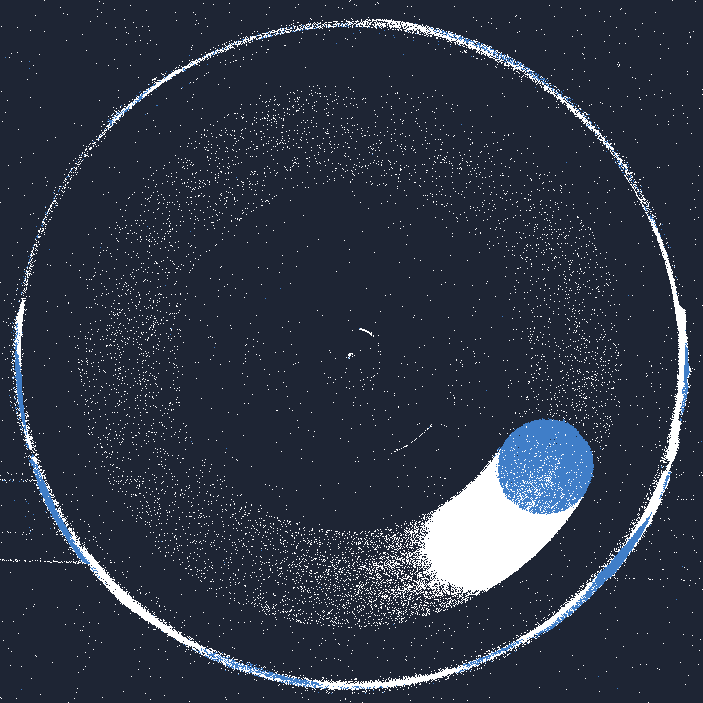}
    \caption{off = 40 and on = -35}
  \end{subfigure}
  \hfill
  \begin{subfigure}{0.19\linewidth}
    \includegraphics[width=\linewidth]{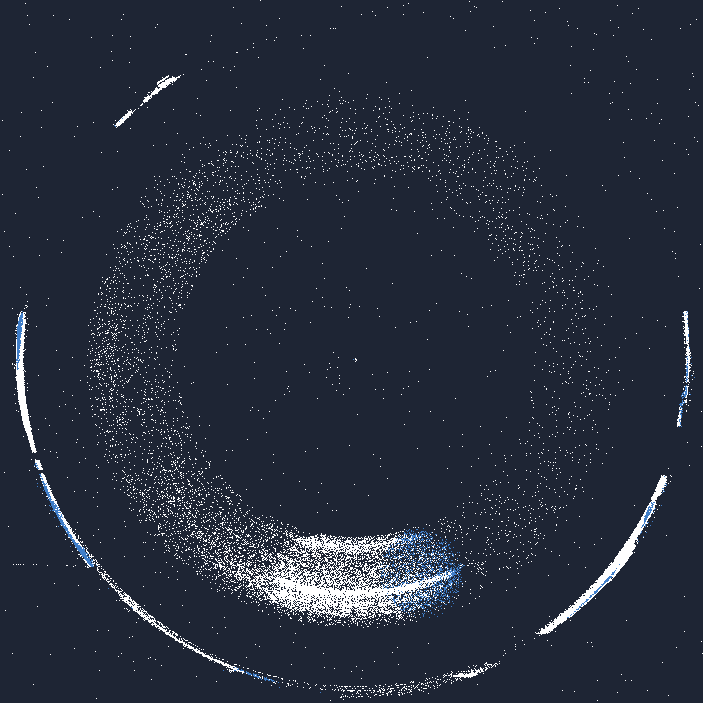}
    \caption{off = 190 and on = -35}
  \end{subfigure}
  \hfill
  \begin{subfigure}{0.19\linewidth}
    \includegraphics[width=\linewidth]{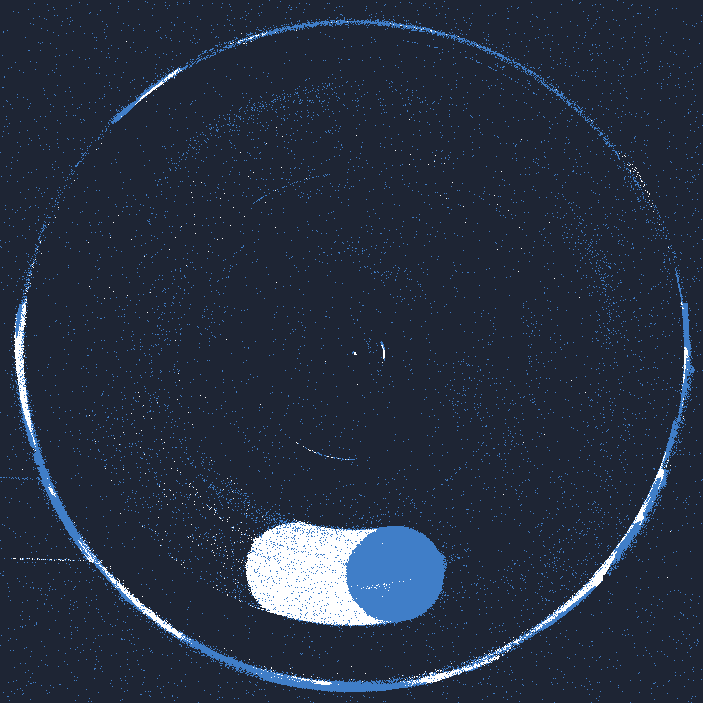}
    \caption{off = -35 and on = 40}
  \end{subfigure}
  \hfill
  \begin{subfigure}{0.19\linewidth}
    \includegraphics[width=\linewidth]{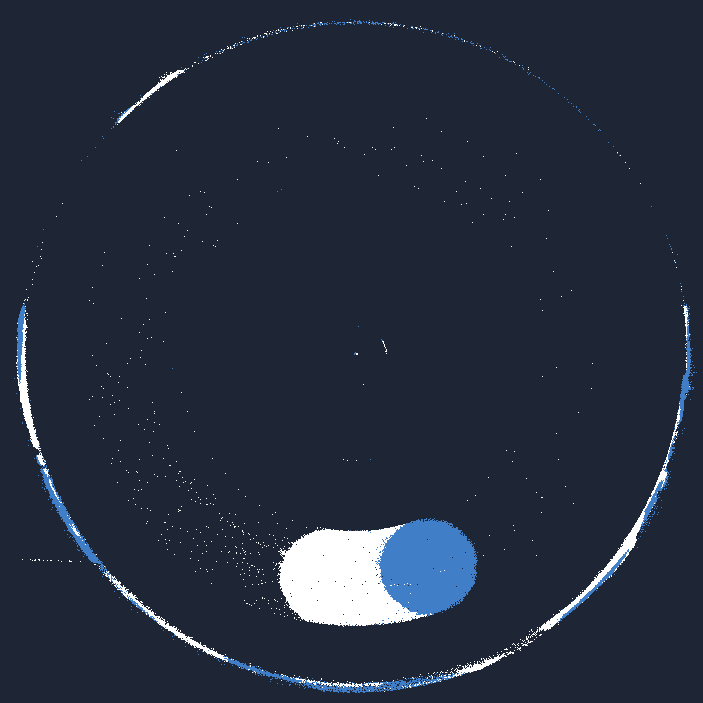}
    \caption{off = 40 and on = 40}
  \end{subfigure}
  \hfill
  \begin{subfigure}{0.19\linewidth}
    \includegraphics[width=\linewidth]{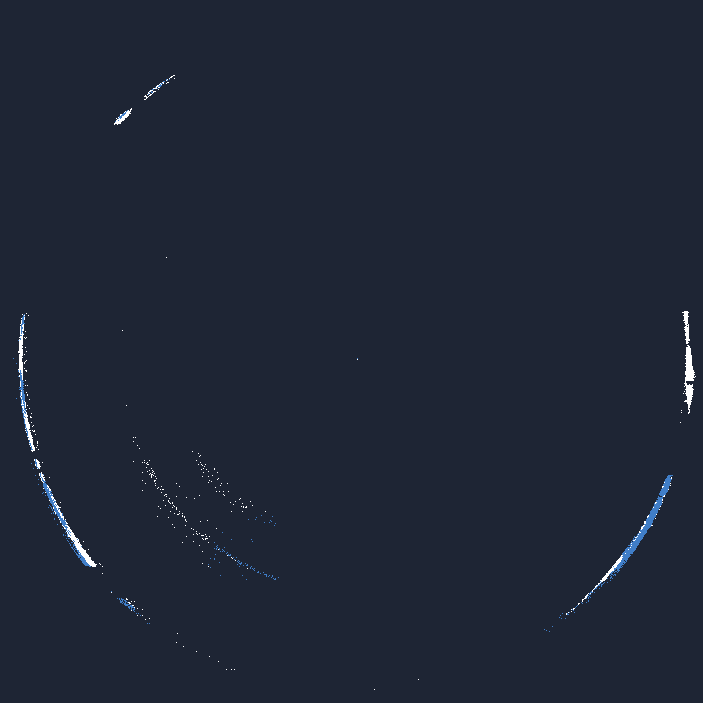}
    \caption{off = 190 and on = 40}
  \end{subfigure}
  \hfill
  \begin{subfigure}{0.19\linewidth}
    \includegraphics[width=\linewidth]{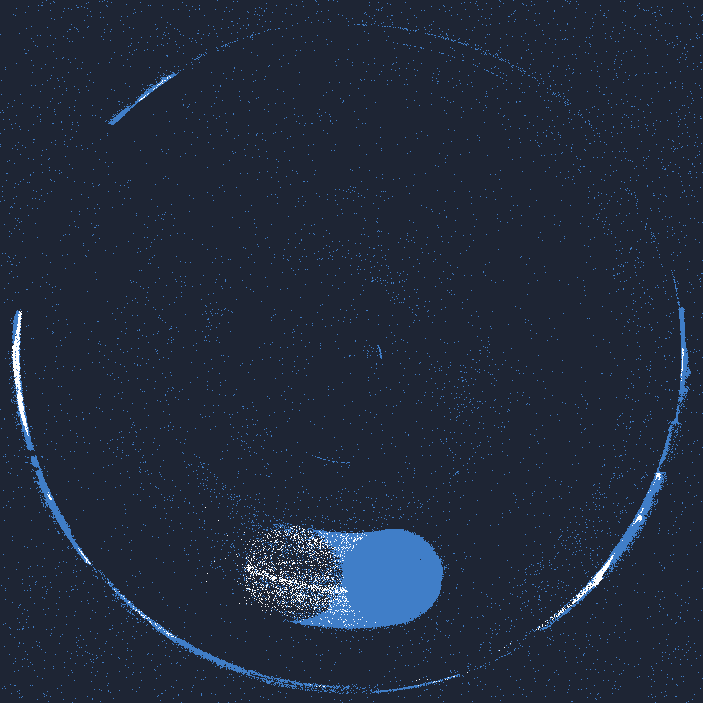}
    \caption{off = -35 and on = 115}
  \end{subfigure}
  \hfill
  \begin{subfigure}{0.19\linewidth}
    \includegraphics[width=\linewidth]{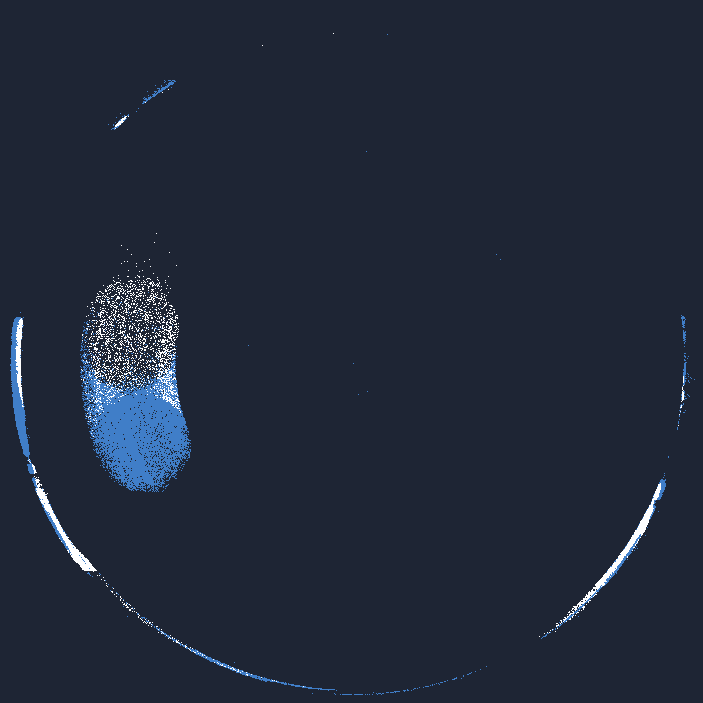}
    \caption{off = 40 and on = 115}
  \end{subfigure}
  \hfill
  \begin{subfigure}{0.19\linewidth}
    \includegraphics[width=\linewidth]{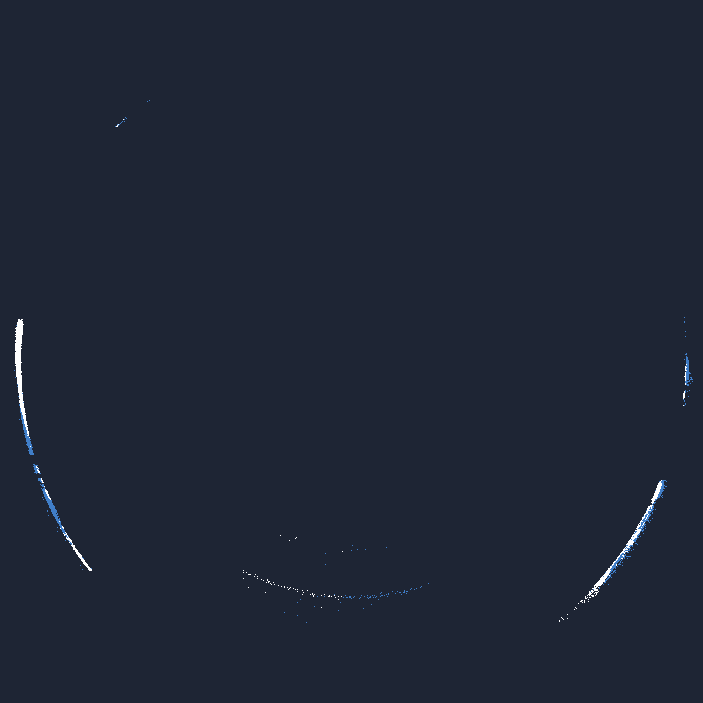}
    \caption{off = 190 and on = 115}
  \end{subfigure}
  \caption{
  A selection of screen captures of Metavision Studio from $100$ recordings of a \textbf{grey dot} with their bias settings, where off refers to \textit{bias\_diff\_off} and on refers to \textit{bias\_diff\_on.}}
  \label{fig:spinning_grey_dot_examples}
\end{figure*}

\begin{figure*}[htb!]
  \centering
  \begin{subfigure}{0.19\linewidth}
    \includegraphics[width=\linewidth]{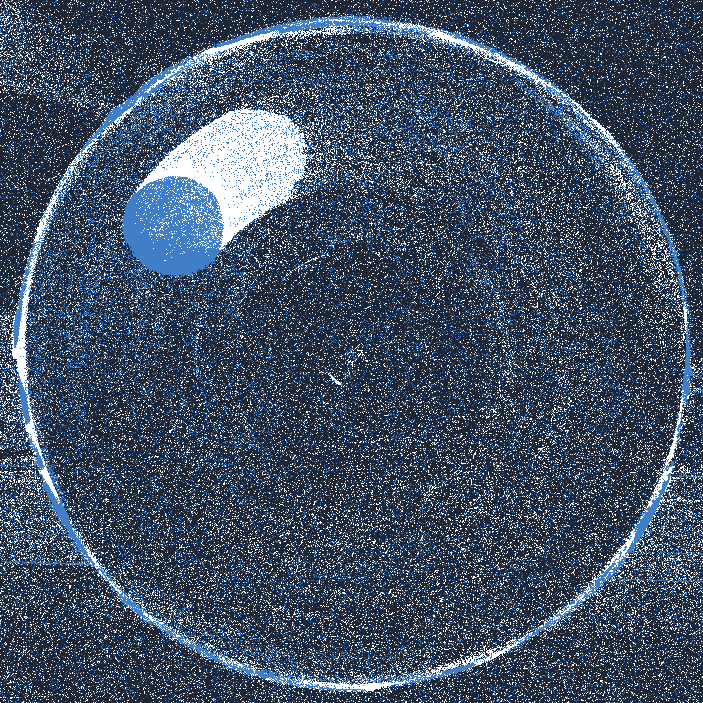}
    \caption{off = -35 and on = -35}
  \end{subfigure}
  \hfill
  \begin{subfigure}{0.19\linewidth}
    \includegraphics[width=\linewidth]{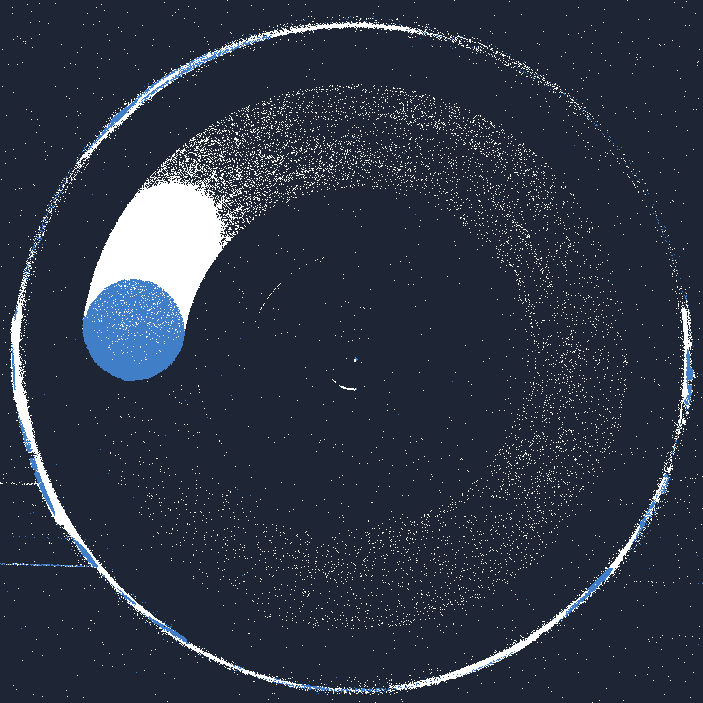}
    \caption{off = 40 and on = -35}
  \end{subfigure}
  \hfill
  \begin{subfigure}{0.19\linewidth}
    \includegraphics[width=\linewidth]{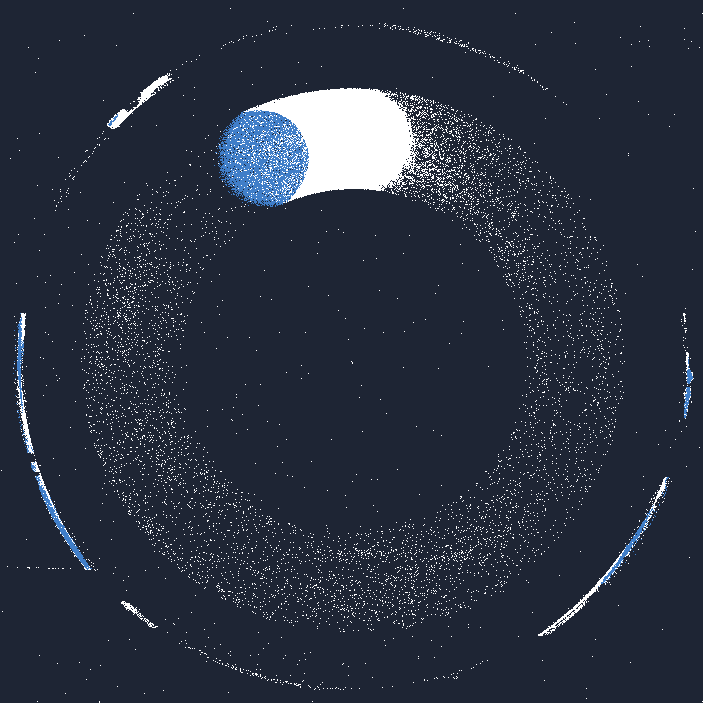}
    \caption{off = 190 and on = -35}
  \end{subfigure}
  \hfill
  \begin{subfigure}{0.19\linewidth}
    \includegraphics[width=\linewidth]{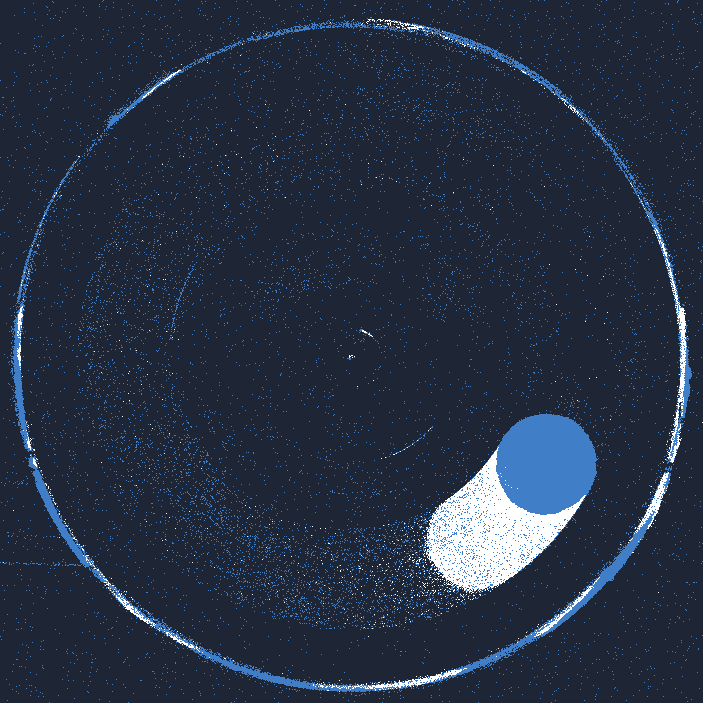}
    \caption{off = -35 and on = 40}
  \end{subfigure}
  \hfill
  \begin{subfigure}{0.19\linewidth}
    \includegraphics[width=\linewidth]{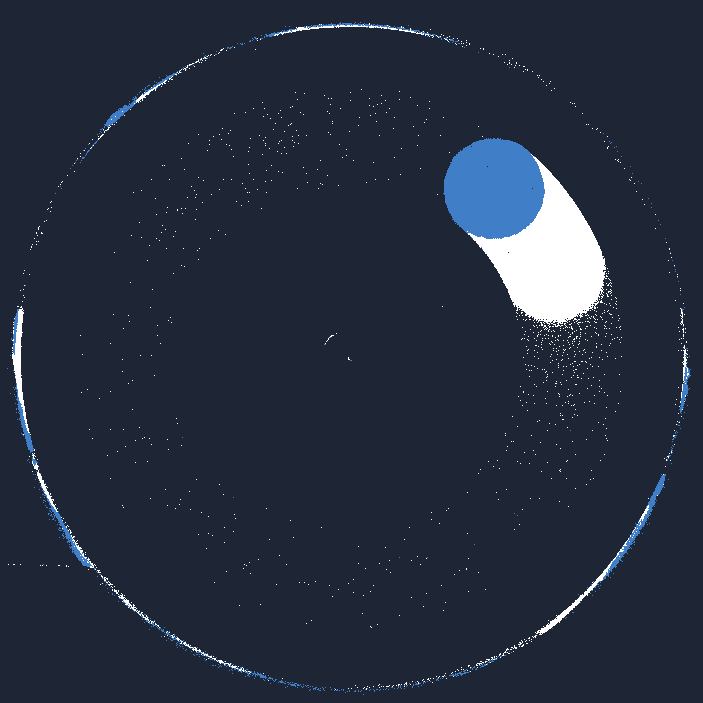}
    \caption{off = 40 and on = 40}
  \end{subfigure}
  \hfill
  \begin{subfigure}{0.19\linewidth}
    \includegraphics[width=\linewidth]{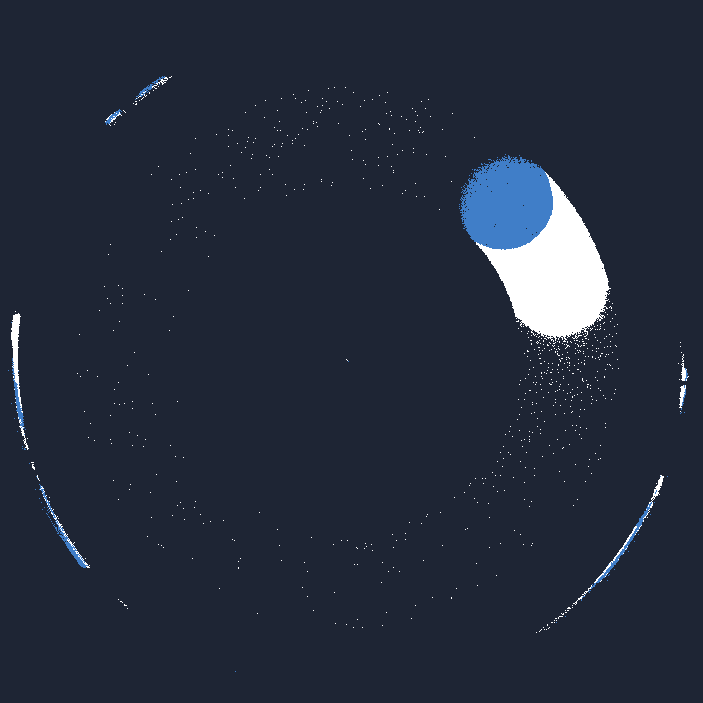}
    \caption{off = 190 and on = 40}
  \end{subfigure}
  \hfill
  \begin{subfigure}{0.19\linewidth}
    \includegraphics[width=\linewidth]{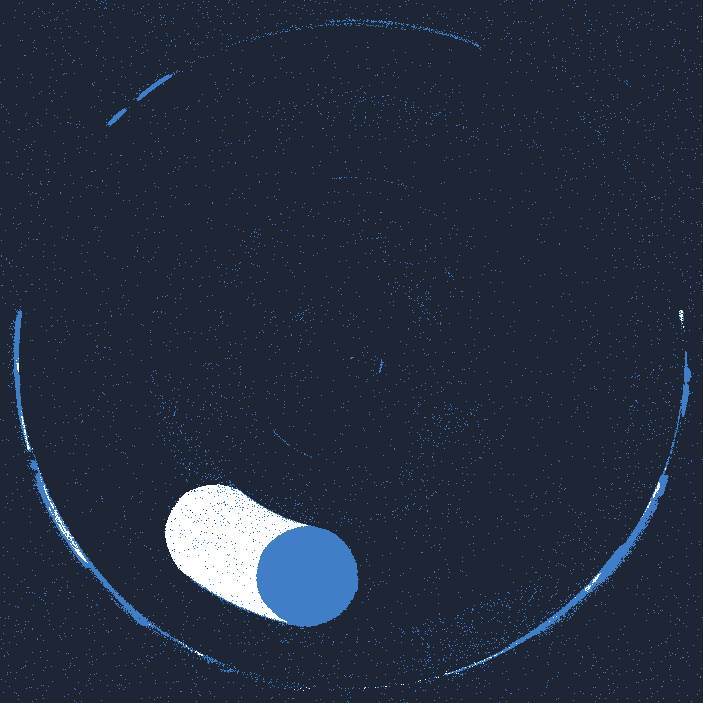}
    \caption{off = -35 and on = 115}
  \end{subfigure}
  \hfill
  \begin{subfigure}{0.19\linewidth}
    \includegraphics[width=\linewidth]{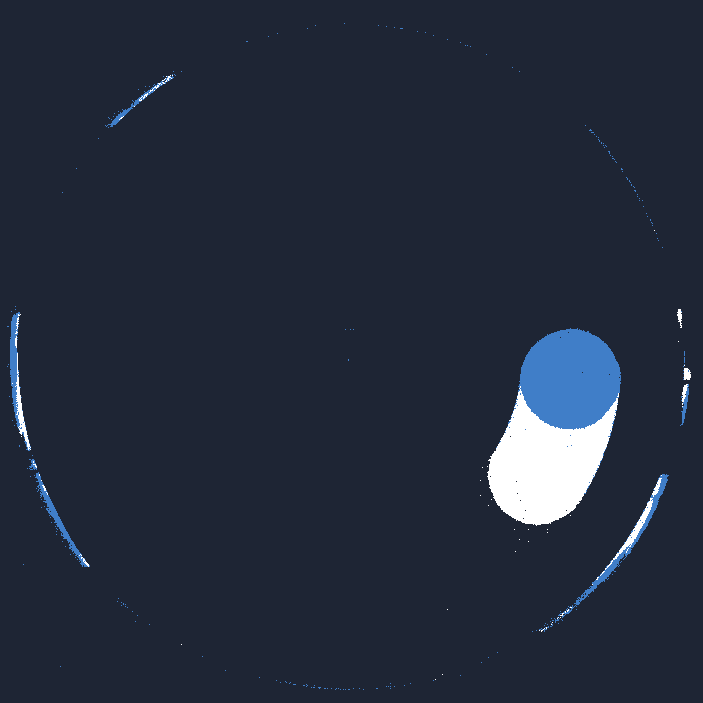}
    \caption{off = 40 and on = 115}
  \end{subfigure}
  \hfill
  \begin{subfigure}{0.19\linewidth}
    \includegraphics[width=\linewidth]{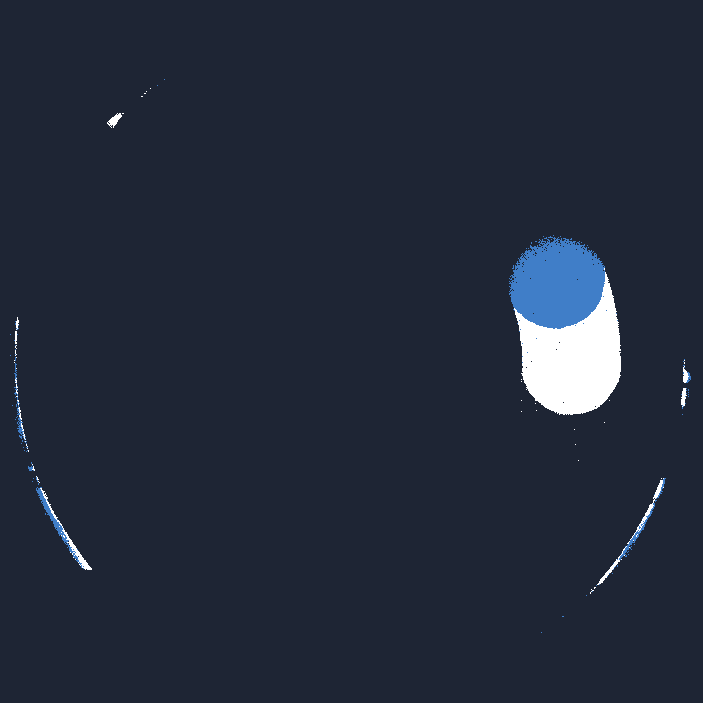}
    \caption{off = 190 and on = 115}
  \end{subfigure}
  \caption{
  A selection of screen captures of Metavision Studio from $100$ recordings of a \textbf{black dot} with their bias settings, where off refers to \textit{bias\_diff\_off} and on refers to \textit{bias\_diff\_on.}}
  \label{fig:spinning_black_dot_examples}
\end{figure*}

The setup of the spinning dot is shown in~\cref{fig:spinning_dot_setup}.
A selection of screen captures from the recordings of the grey dot with their bias settings is visualized in~\cref{fig:spinning_grey_dot_examples}.
And for the black dot with their bias settings in~\cref{fig:spinning_black_dot_examples}.

\subsection{Blinking LED Board}\label{subsec:sm_led_board}

\begin{figure}[t!]
  \centering
  \includegraphics[width=1.0\linewidth]{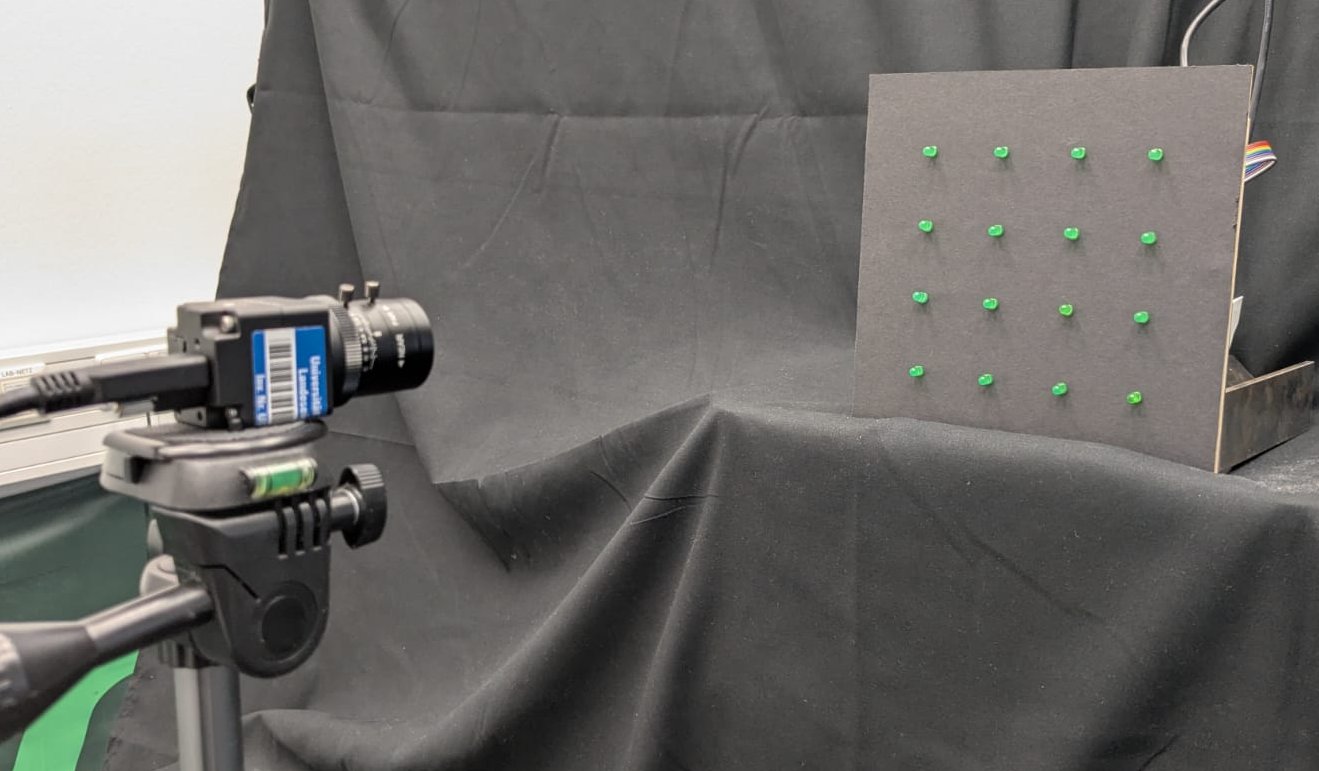}
  \caption{
    A blinking LED board consisting of LEDs with different frequencies and waveforms.
  }
  \label{fig:led_board_setup}
\end{figure}

A blinking LED board, shown in~\cref{fig:led_board_setup}, was specifically designed as an evaluation tool for event cameras in a scientific study.
Its purpose is to enable the simultaneous observation of varying frequencies and luminosity gradients, offering a robust testing environment for evaluating event camera performance under diverse conditions.

The board consists of a total of 16 LEDs, categorized as follows:

\begin{itemize}
    \item \textbf{High-Frequency Blinking LEDs (4 LEDs)}: These LEDs blink at high frequencies that are not perceivable by the human eye (\textgreater 50 Hz). They are particularly sensitive to the \textit{bias\_hpf} bias setting, as higher cutoff frequencies in the high-pass filter will predominantly retain events from these high-frequency LEDs while filtering out lower-frequency signals.

    \item \textbf{Sinusoidal Luminosity LEDs (6 LEDs)}: These LEDs modulate their brightness following a sinusoidal wave at different frequencies. The smooth, continuous change in luminosity creates a distinct time gradient in the events generated by the camera.

    \item \textbf{Triangular Wave Luminosity LEDs (6 LEDs)}: These LEDs also change their brightness at the same set of frequencies as the sinusoidal LEDs, but with a triangular wave pattern. The sharper transitions of the triangular wave offer a different type of temporal gradient for evaluation.
\end{itemize}

The difference in frequencies across the sinusoidal and triangular LEDs results in varying temporal gradients, allowing for a nuanced assessment of how the event camera responds to different rates of change in light intensity.

The board's design also allows for testing bias tuning algorithms:

\begin{itemize}
    \item Adjusting the \textit{bias\_diff\_on} and \textit{bias\_diff\_off} parameters can suppress the events generated by low-frequency LEDs, effectively isolating high-frequency activity.
    
    \item The \textit{bias\_fo} setting provides additional filtering by attenuating signals from lower frequency LEDs, enabling more precise control over the frequency response of the camera.
\end{itemize}

The LED board is controlled by a microcontroller, allowing easy adjustment of LED frequencies. 
While Pulse Width Modulation (PWM) is commonly used for LED intensity control, the event camera would detect the rapid on-off transitions. 
To avoid this, capacitors were added in parallel to the sinusoidal and triangular LEDs, acting as integrators to smooth the PWM signal into a continuous analog waveform. 
This ensures gradual brightness changes and prevents unwanted high-frequency noise.

We recorded $30,976$ sequences, covering the following biases:
\begin{itemize}
    \item \textit{bias\_diff\_on}: -80, -60, -40, -20, 0, 20, 40, 60, 80, 100, 120
    \item \textit{bias\_diff\_off}: -30,-10, 10, 30, 50, 70, 90, 110, 130, 150, 170
    \item \textit{refr\_time}: 4, 20, 36, 52, 68, 84, 100, 116
    \item \textit{refr\_fo}: -29, -18, -7, 4, 15, 26, 37, 48
    \item \textit{refr\_hpf}: -15, 65, 145, 225
\end{itemize}

\begin{figure}[htb!]
  \centering
  \includegraphics[width=1.0\linewidth]{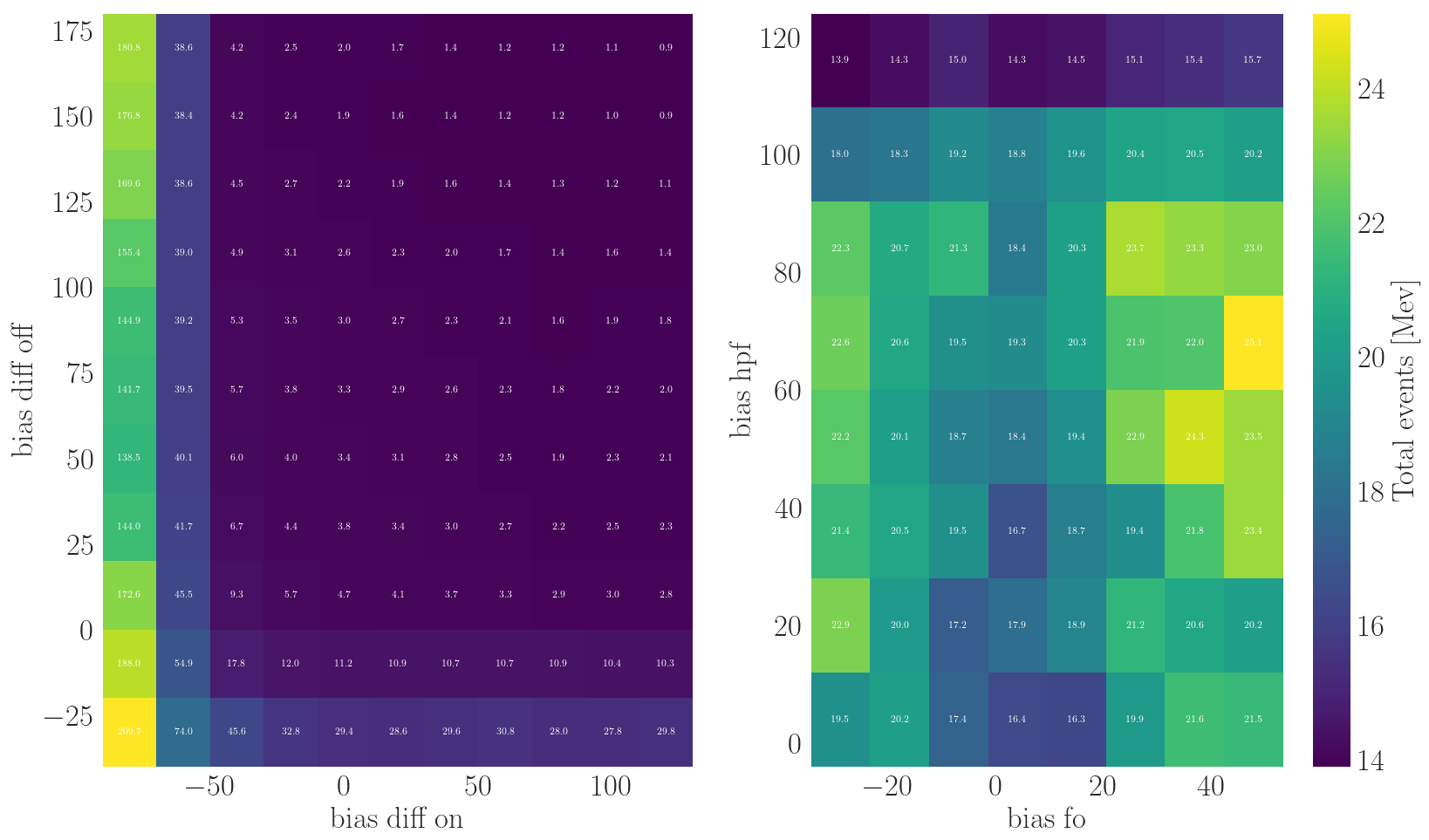}
  \caption{
    The average event Rate for different bais combinations.
    \textit{bias\_diff\_off} vs \textit{bias\_diff\_on} on the left side, \textit{bias\_hpf} vs \textit{bioas\_fo} on the right side. 
    The lowest threshold bins contain roughly $50\%$ of all events produced.
  }
  \label{fig:event_rate_board}
\end{figure}

\begin{figure}[htb!]
  \centering
  \includegraphics[width=1.0\linewidth]{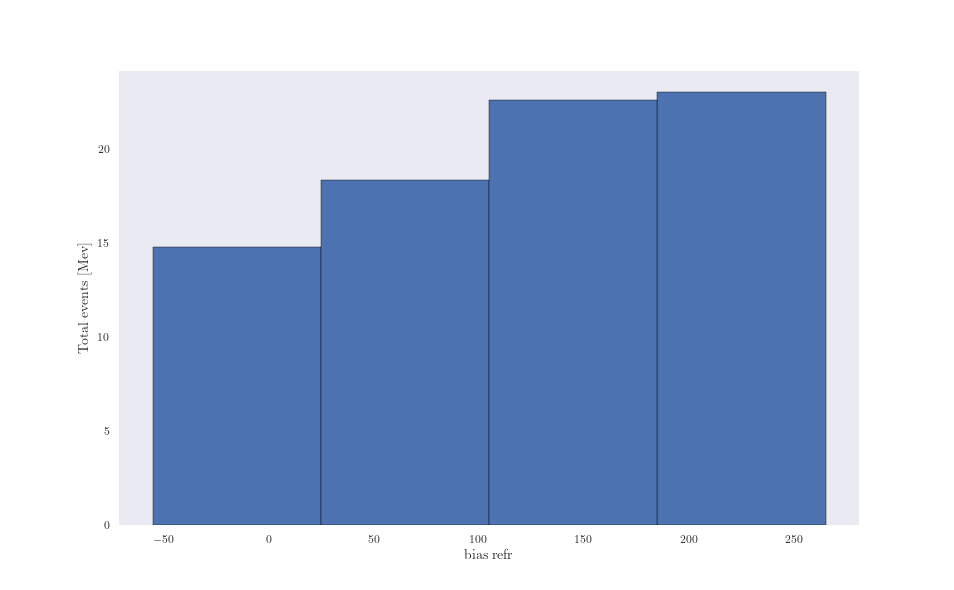}
  \caption{
    The average event Rate for different refactory periods.
  }
  \label{fig:ref_board}
\end{figure}

\Cref{fig:event_rate_board} shows the average \ac{ER} over the whole dataset for each bias setting.
The bias combinations with the lowest value for one of the two thresholds contain around $50\%$ of the total event count.
Additionally, a very high high-pass filter can be used to get rid of many events.
A decrease in the refactory period also decreases the average event rate as displayed in \Cref{fig:ref_board}.

\begin{figure}[t!]
  \centering
  \includegraphics[width=1.0\linewidth]{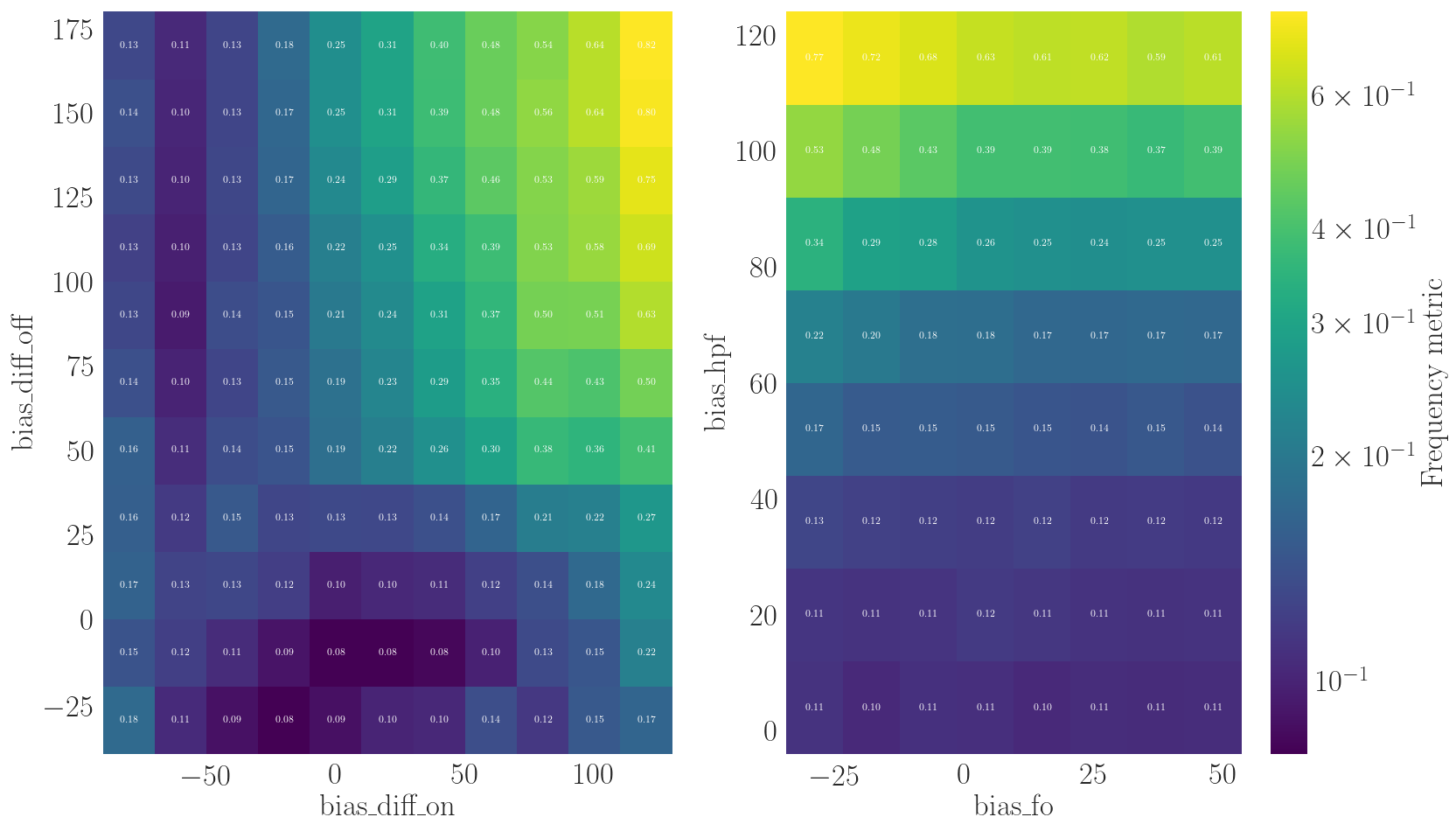}
  \caption{
    The average RFU for different bias combinations
    \textit{bias\_diff\_off} vs \textit{bias\_diff\_on} on the left side, \textit{bias\_hpf} vs \textit{bias\_fo} on the right side. 
    The most efficient frequency estimation can be achieved with a $\textit{bias\_diff\_off} = -10$, $\textit{bias\_diff\_on} = 20$ and $\textit{bias\_hpf} = 0$.
  }
  \label{fig:board_RFU}
\end{figure}

The slow-blinking LEDs start to vanish for high bias thresholds.
Resulting in an increased RFU in these regions.
This can be seen in \cref{fig:board_RFU}.
The increase of noise in low bias setting also hinders the frequency detection, therefore, an optimal around $\textit{bias\_diff\_off} =-20$, $\textit{bias\_diff\_on} = 20$ results in the highest average RFU.
A strong high-pass filter in combination with a strong low-pass filter also decreases the sensitivity for the detection of some LEDs, therefore, the RFU also increases in these regions significantly.

\subsection{Visual Odometry (VO)}\label{subsec:sm_vo}

\begin{figure}[ht!]
  \centering
  \includegraphics[width=1.0\linewidth]{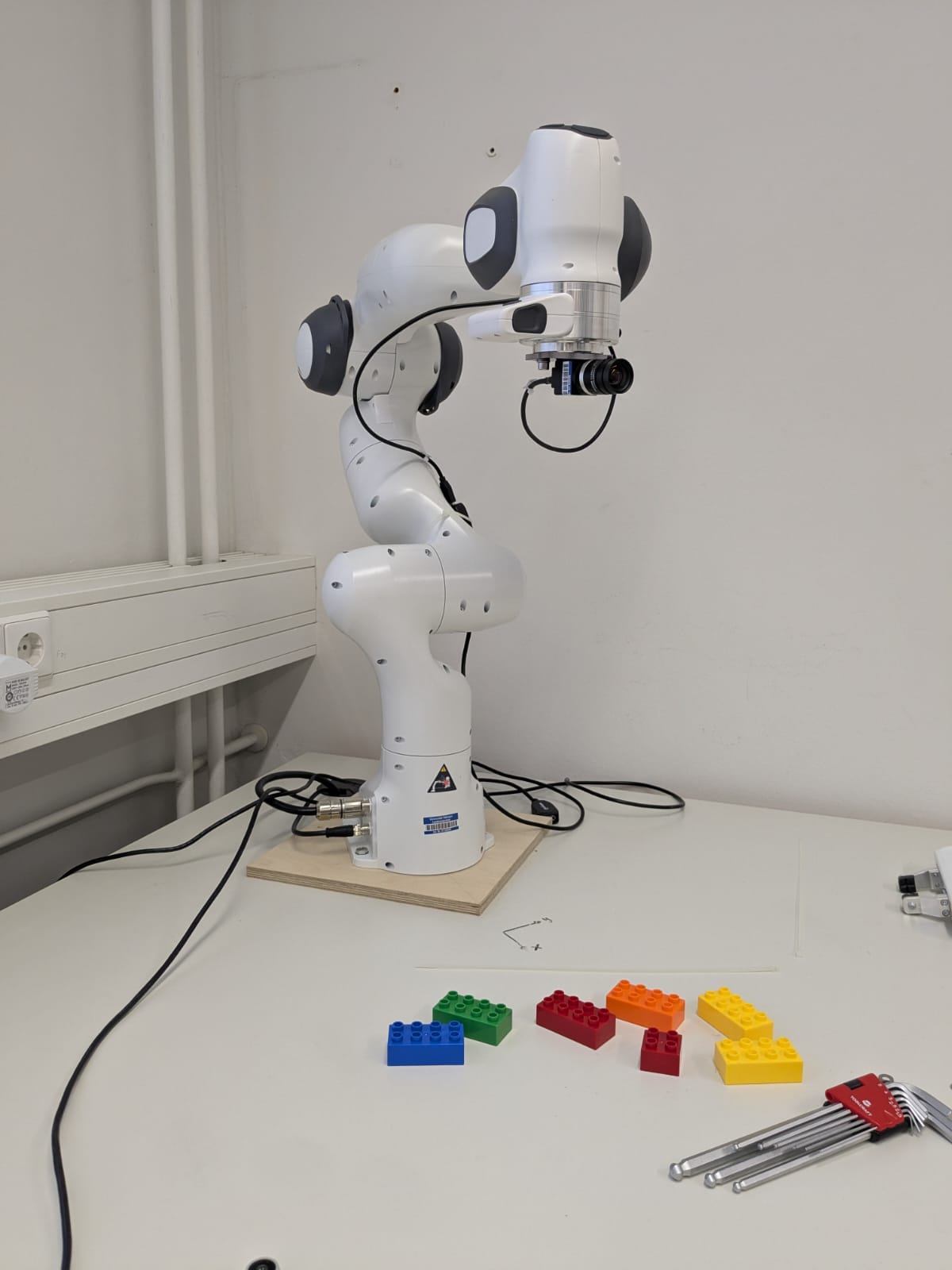}
  \caption{
    The setup to record the VO dataset.
  }
  \label{fig:vo_setup}
\end{figure}

To get a repeatable motion and an accurate ground truth trajectory, we mounted an event camera to the end-effector of a Pandas robot arm, shown in~\cref{fig:vo_setup}, to move the event camera along the same trajectory with different biases.
The scene was a static environment under constant illumination.

\begin{figure}[ht!]
  \centering
  \includegraphics[width=1.0\linewidth]{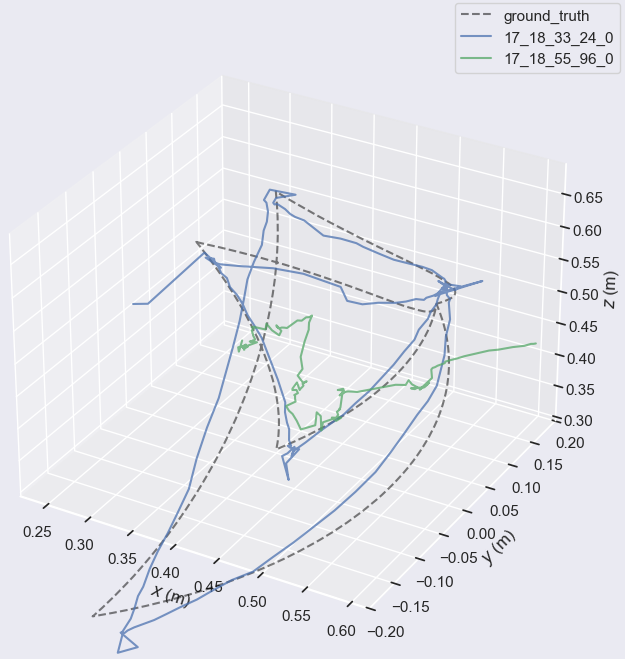}
  \caption{
    The ground truth trajectory of the end-effector of a Pandas robot arm is shown in {\color{black} in black}.
    The trajectory output from DEVO~\cite{Klenk20243dv} for the \textit{bias} settings $(17, 18, 33, 24)$ is shown in {\color{blue} in blue}.
    And the trajectory output for the \textit{bias} settings $(17, 18, 55, 96)$ is shown in {\color{green} in green}.
    Despite the similar bias values, the \ac{VO} failed in one case.
  }
  \label{fig:vo_triangle_trajectory}
\end{figure}

The second ground truth trajectory from the end-effector of the robot arm is shown in~\cref{fig:vo_triangle_trajectory}.
Although the bias settings of the two trajectories shown in~\cref{fig:vo_triangle_trajectory} are similar, DEVO successfully inferred the trajectory in one case but failed in the other.
As previously mentioned, this underscores the fact that even when the event stream looks comparable to human observers, its effectiveness in downstream tasks can vary significantly.

\begin{figure}[t!]
  \centering
  \includegraphics[width=1.0\linewidth]{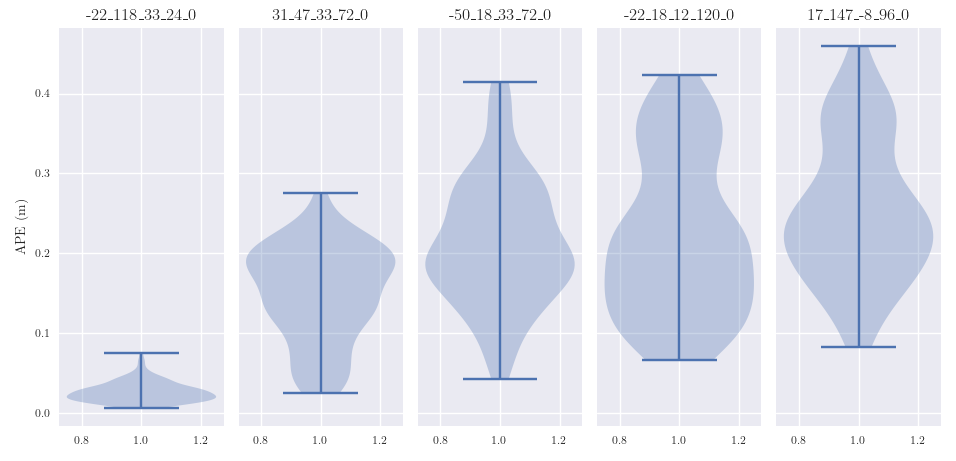}
  \caption{
        The \acf{APE} plot of five trajectories, each with different bias settings, estimated by DEVO~\cite{Klenk20243dv}.
        These five trajectories are picked to have an equidistant error between each other.
        }
  \label{fig:vo_triangle_errors}
\end{figure}

From all the recordings of the second trajectory, we visualize the \ac{APE} of five selected ones in~\cref{fig:vo_triangle_errors}.
These recordings are chosen such that their errors are evenly spaced across the range of observed errors.

From~\cref{fig:vo_triangle_errors}, it can be seen that only the recording with the bias settings $(-22, 118, 33, 24, 0)$ led to a reasonable \ac{APE}.
This shows the importance of bias tuning in order for a downstream application to give reasonable results.

We consider biases valid when $\textit{APE} < 0.13$.
Under this criterion, $18.81\%$ of the settings result in a successful trajectory estimate.

\section{Baseline}\label{sec:sm_baseline}

\begin{figure}[ht!]
  \centering
  \includegraphics[trim={0.6cm 0.4cm 0.6cm 0.4cm},clip,width=1.0\linewidth]{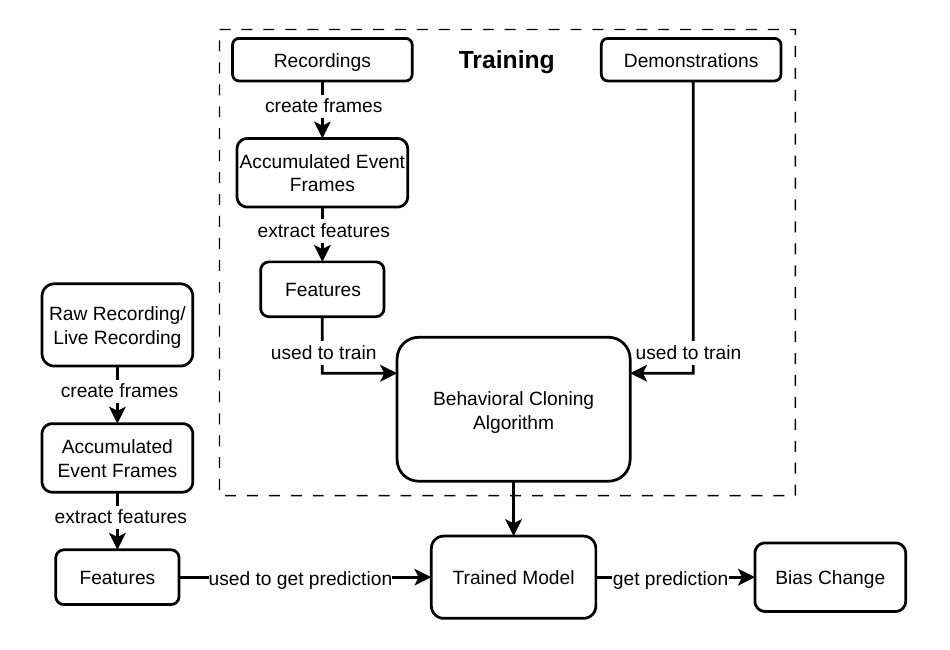}
  \caption{
    An overview of the \acf{BC} pipeline.
  }
  \label{fig:training_pipeline}
\end{figure}

The overall pipeline used for this learning process is illustrated in~\cref{fig:training_pipeline}.

In order to rate the output of the event camera, the primary metric we chose was whether the dot on an accumulated event frame is visible to the human eye or not.
This is the first and foremost concern because if the dot is not visible, it defeats the entire point of the recording.
The secondary metric that we chose is the \ac{ER}, which should be as low as possible while keeping the signal strong.
This translates to low background noise.
We define an optimum bias setting as any setting that fulfills the two conditions.

\begin{figure}[htb!]
  \centering
  \includegraphics[width=1.0\linewidth]{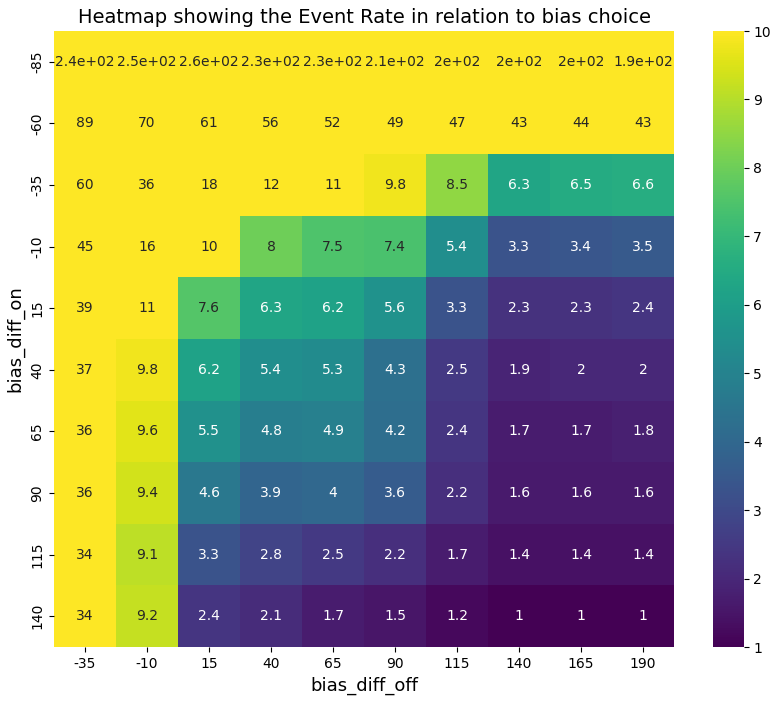}
  \caption{
    A heatmap showing the \acf{ER} for different bias choices of the spinning \textbf{grey} dot recording.
  }
  \label{fig:event_rates_grey_dot}
\end{figure}

\begin{figure}[htb!]
  \centering
  \includegraphics[width=1.0\linewidth]{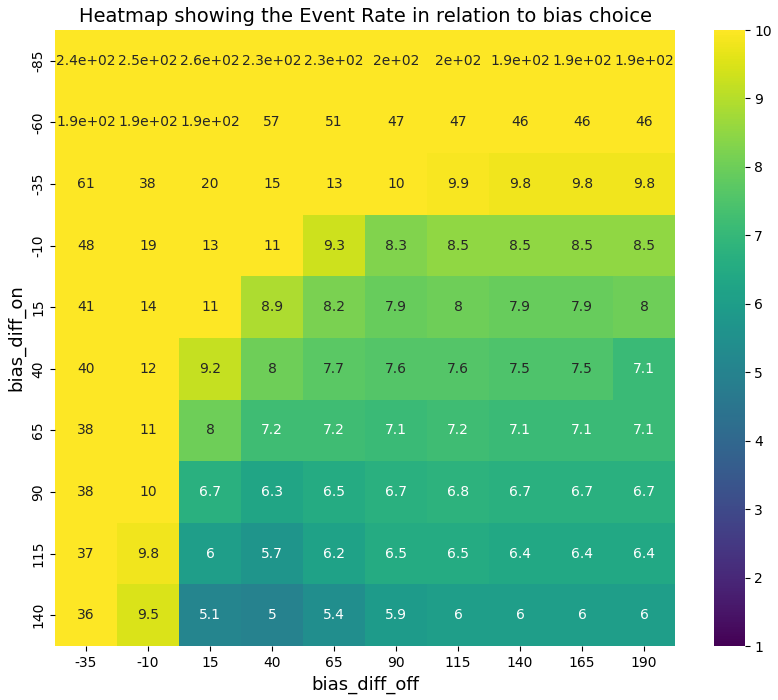}
  \caption{
    A heatmap showing the \acf{ER} for different bias choices of the spinning \textbf{black} dot recording.
  }
  \label{fig:event_rates_black_dot}
\end{figure}

The average \ac{ER} was calculated for each recording by dividing the total amount of events by the length of the recording.
The \ac{ER} is used to measure the amount of signal and noise and to try to find a bias setting with a good ratio between the two.
As can be observed in~\cref{fig:event_rates_grey_dot} and in~\cref{fig:event_rates_black_dot}, for recordings where either of the two biases has a small value, the \ac{ER} is larger, by order of magnitude, compared to the recordings where the biases had higher values.
The recordings with the \textit{bias\_diff\_on} values between $-85$ and $-60$ contain errors because the readout bus of the event camera did not have enough bandwidth to transmit all the events.
This means that the \ac{ER} is bigger than the maximum allowed one.

The most noticeable difference between the heatmap of the \ac{ER} for the grey dot in~\cref{fig:event_rates_grey_dot} and the one for the black dot in~\cref{fig:event_rates_black_dot} is that the average \ac{ER} is higher for the black one.
This is expected, as the contrast between the black dot and the white background is higher than for the grey dot.

To find the range of the optimal biases, we also look at the accumulated event frames.
An accumulation period of $8$ ms has been chosen so that the entire body of the dot can be seen.
The first thing noticeable is that for high values of \textit{bias\_diff\_off}, the dot is no longer visible, as seen in \cref{fig:spinning_grey_dot_examples}.
The camera has such a low sensitivity for those settings that the contrast between the dot and the background is no longer enough to produce any events.
As we observed from the \ac{ER}, for too small values of both biases, especially if they are negative, the noise severely affects the quality of the recording, even if the dot is visible.

For the grey dot, an optimal bias, as we defined it, is $\textit{bias\_diff\_off} = 40$ and $\textit{bias\_diff\_on} = 40$.
For this setting, the dot is visible, and the noise is greatly reduced.
We have considered that for the setting $\textit{bias\_diff\_off} = 40$ and $\textit{bias\_diff\_on} = 115$, the signal was too reduced for it to be considered an optimal bias, even though the dot is visible to the human eye.
The observations made from viewing the recordings helped us conclude that optimal biases can be found in the range $[15, 65]$ for \textit{bias\_diff\_off} and in the range $[40, 90]$ for \textit{bias\_ diff\_on}.

For the black dot, we can observe that the dot is visible even for the maximum values of the thresholds, while the grey one disappears.
Because of this, the range of the optimal biases is larger than for the grey dot, being $[40, 165]$ for \textit{bias\_diff\_off} and $[40, 115]$ for \textit{bias\_diff\_on}.

We have chosen to define the action in this environment as the relative bias change, expressed with an integer to be added to the current bias value, as this is the way we went about tuning the biases by looking at the output and approximating how big of a change is necessary.
Thus, we have written demonstrations consisting of a tuple of actions with the form $(a, b, 0, 0, 0)$, where $a$ represents the change of \textit{bias\_diff\_off} and $b$ represents the change of \textit{bias\_diff\_on} that is needed to reach the range of optimal biases.

\paragraph{Different Feature Extractors}

As an ablation study for the feature extraction, we have adapted two models of ResNet: ResNet18, with $18$ layers, and ResNet50, with $50$ layers, a more complex network, capable of capturing more information.
The choice of the network has an influence on the ability of the network to learn.

ResNet50 is a more complex and deep network compared to ResNet18, and therefore, the \ac{BC} model is able to finish the learning process in $1000$ steps.
Due to its more complex structure, ResNet50 creates a more information-dense latent space.
The downside of this is a more computationally expensive training process for the model, which was already done on ImageNet and is of no concern for this work.
The exact loss values after the training process finished are listed in~\cref{tab:resnet}.

As evident from the training and test loss in~\cref{tab:resnet}, ResNet50 is the better option for our purposes, and therefore, we decided to use ResNet50 as our feature extractor.

\begin{table}[htb!]
\begin{center}
    \begin{tabular}{ll}
      \textbf{Model} & \textbf{Test loss} \\
      \hline
      \hline
      ResNet18 & 0.24\\
      \hline
      ResNet50 & 0.2\\
      \hline
    \end{tabular}
    \caption{
        The final loss values when the training process is finished.
        The loss is displayed as the Mean Squared Error of the normalized proposed action.}
  \label{tab:resnet}
\end{center}
\end{table}

\paragraph{Different Accumulation Times}

As described in~\cref{subsec:method}, we are using accumulated event frames to extract features from the event stream.
In an additional ablation study, we investigated the influence of the accumulation time.
Depending on the application, the duration of the accumulation time can have a significant influence on the performance.
Therefore, we investigated the influence of different accumulation times ($1$ ms, $2$ ms, $8$ ms, and $16$ ms).
\begin{table}[htb!]
\begin{center}
    \begin{tabular}{rll}
        \textbf{Accumulation Time} & \textbf{Test loss} \\
        \hline
        \hline
        1 ms & 0.2\\
        \hline
        2 ms & 0.22\\
        \hline
        8 ms & 0.21\\
        \hline
        16 ms & 0.27\\
        \hline
    \end{tabular}
    \caption{
        The final loss values with different accumulation times after the training process finished.
        The loss is given in the mean squared Error of the normalized proposed action.}
    \label{tab:acc_time}
\end{center}
\end{table}

From the final test losses in~\cref{tab:acc_time}, we can see that for our data, a shorter accumulation time is favorable.

\end{document}